\documentclass[bst/sn-mathphys,Numbered]{sn-jnl}

\usepackage{geometry}

\newgeometry{vmargin={25mm}, hmargin={35mm,35mm}}   

\usepackage{graphicx}%
\usepackage{multirow}%
\usepackage{amsmath,amssymb,amsfonts}%
\usepackage{amsthm}%
\usepackage{bbold}%
\usepackage{mathrsfs}%
\UseRawInputEncoding
\usepackage[title]{appendix}%
\usepackage{xcolor}%
\usepackage{textcomp}%
\usepackage{manyfoot}%
\usepackage{booktabs}%
\usepackage{algorithm}%
\usepackage{algorithmic}%
\usepackage{listings}%
\usepackage{makecell}%
\usepackage{float}%
\usepackage{caption}%
\usepackage{subcaption}%
\usepackage{soul}%

\theoremstyle{thmstyleone}%
\theoremstyle{thmstyletwo}%

\theoremstyle{thmstylethree}%

\begin{document}

\title[Article Title]{Deep Trees for (Un)structured Data: Tractability, Performance, and Interpretability}


\author*[1,2]{\fnm{Dimitris} \sur{Bertsimas}}\email{dbertsim@mit.edu}

\author[1]{\fnm{Lisa} \sur{Everest}}

\author[1]{\fnm{Jiayi} \sur{Gu}}

\author[1]{\fnm{Matthew} \sur{Peroni}}

\author[1]{\fnm{Vasiliki} \sur{Stoumpou}}

\affil*[1]{\orgdiv{Operations Research Center}, \orgname{Massachusetts Institute of Technology}, \orgaddress{\city{Cambridge}, \state{MA}, \country{USA}}}

\affil[2]{\orgdiv{Sloan School of Management}, \orgname{Massachusetts Institute of Technology}, \orgaddress{\city{Cambridge}, \state{MA}, \country{USA}}}

\abstract{Decision Trees have remained a popular machine learning method for tabular datasets, mainly due to their interpretability. However, they lack the expressiveness needed to handle highly nonlinear or unstructured datasets. Motivated by recent advances in tree-based machine learning (ML) techniques and first-order optimization methods, we introduce Generalized Soft Trees (GSTs), which extend soft decision trees (STs) and are capable of processing images directly. We demonstrate their advantages with respect to tractability, performance, and interpretability. We develop a \textit{\textbf{tractable}} approach to growing GSTs, given by the DeepTree algorithm, which, in addition to new regularization terms, produces high-quality models with far fewer nodes and greater interpretability than traditional soft trees. We test the \textit{\textbf{performance}} of our GSTs on benchmark tabular and image datasets, including MIMIC-IV, MNIST, Fashion MNIST, CIFAR-10 and Celeb-A. We show that our approach outperforms other popular tree methods (CART, Random Forests, XGBoost) in almost all of the datasets, with Convolutional Trees having a significant edge in the hardest CIFAR-10 and Fashion MNIST datasets. Finally, we explore the \textit{\textbf{interpretability}} of our GSTs and find that even the most complex GSTs are considerably more interpretable than deep neural networks. Overall, our approach of Generalized Soft Trees provides a tractable method that is high-performing on (un)structured datasets and preserves interpretability more than traditional deep learning methods.}




\maketitle

\section{Introduction}\label{sec:intro}

The rapid developments in deep learning algorithms, access to a variety of data sources, and increase in computational power in the past years have resulted in the explosion of Deep Learning models, which have been tremendously successful in handling multimodal data. However, this success comes at a cost. Despite recent efforts to provide some degree of interpretability \cite{chattopadhyay_grad-cam_2018, feature_vis,lundberg_unified_2017}, deep neural networks remain largely black-box models \cite{slack_fooling_2020,subramanya_fooling_2019}. On the other hand, traditional tree-based methods, although they have less expressive power and cannot handle multimodal data, still remain relevant for tabular datasets, since they achieve strong performance and, most importantly, are fully interpretable. Tree ensembles, like Random Forests \cite{random-forest} and Boosted Trees \cite{xgboost} are not interpretable, but they have strong performance on tabular data, often outperforming deep learning approaches.  

In this paper, we combine the advantages of the two aforementioned methodologies and introduce Generalized Soft Trees (GSTs), an extension of regular Soft Trees. Soft Trees refer to tree models, in which a sample is not solely assigned to a specific leaf node, but rather has a soft assignment to multiple leaves. Our GSTs preserve this property but are generalized in the sense that they contain more general functions in their nodes. Specifically, we consider both hyperplane splits as well as convolutional splits, which are similar to a single convolutional layer in a neural network.

GSTs are tractable models that leverage features from both Soft Decision Trees and Neural Networks. They are trained using backpropagation and stochastic gradient descent, which makes their training process more efficient than recent optimal tree approaches, such as Optimal Classification Trees \cite{optimal-classification-trees}, scaling to significantly greater tree depth and larger feature spaces. We also present the DeepTree algorithm, a scalable heuristic approach to iteratively growing GSTs by selecting, training, and adding new splits based on potential loss reduction. The algorithm demonstrates that growing GSTs can often efficiently improve performance with the addition of just a few nodes, compared to full trees. 

GSTs also display better performance than other tree methods in a  variety of datasets. One of the most contributing factors is their ability for direct and nonlinear processing of images, an unstructured data modality. Such processing allows GSTs to handle image data without treating it as structured data or using embeddings in a latent feature space. Further, this enables GSTs to perform more advanced feature extraction with enhanced modeling power. We demonstrate this modeling power in benchmark experiments including MIMIC-IV, MNIST, Fashion MNIST, CIFAR-10 and Celeb-A, which span across multiple modalities, such as tabular data and images. We show how GSTs perform either significantly or reasonably well for the majority of our datasets. For CIFAR-10, the hardest image dataset we experimented with, we manage to significantly outperform the rest of the tree models, but our performance is not comparable to deep learning methods. The main reason for that is the multiclassification nature of the dataset, combined with the low image quality, a setting that apparently requires feature extraction from nested convolutional layers to yield good results. When we convert the problem to a binary classification one, the results we get are significantly improved.

We further demonstrate how GSTs provide greater interpretability. Though employing more complex functions in Convolutional Trees reduces interpretability as compared to Hyperplane Trees, we still maintain interpretability in both submodels of GSTs. Using a regularization approach, we aim to resolve an issue commonly encountered in other soft trees, which is the distribution of a specific sample across different leaves. This method limits the number of leaf nodes each sample is assigned to, thus improving our understanding of the classes each leaf node mainly corresponds to. The second way to recover interpretability is by leveraging the coefficients of the tree nodes, and a third one is employing feature visualization techniques. Our GSTs therefore maintain many characteristics of interpretability that are
missing from large models like Convolutional Neural Networks and Transformers.

In summary, we contribute GSTs which are tractable, performant, and considerably interpretable models. We note that our results are attained by training the trees from scratch, without employing knowledge distillation techniques that are used in other related work (\cite{frosst_distilling_2017}).

We present all of our findings in the following sections. Section \ref{sec:related-work} describes related work. Sections \ref{sec:methods} and \ref{deeptree-algo} present our tractable methodology of modeling and growing GSTs using the DeepTree algorithm respectively. Section \ref{sec:results} reports the results of our experiments across different datasets featuring various data modalities. Section \ref{sec:interpret} explores the interpretability of our GST models. Finally, Section \ref{sec:conclusions} contains our discussion and conclusions on this work.

\subsection{Related Work}\label{sec:related-work}
The class of Soft Tree (ST) models that we base this work on was first introduced by \cite{irsoy-soft-2012}. In addition to establishing the notion of trees with soft splits, \cite{irsoy-soft-2012} also introduces the notion of more general nonlinear split functions. However, their work does not actually explore any nonlinear split functions. They also only consider constant leaf nodes, whereas we consider linear leaf nodes. Additionally, they evaluate their models on tabular datasets only and do not consider unstructured data. This work also develops a method for growing soft trees. In their approach, they initialize a tree as a single node and grow the left and right subtrees recursively, training each new split on the training set using backpropagation, until the change in validation error stops decreasing. Our method for growing soft trees, while similar to this idea, uses a heuristic approach to select leaf nodes to split on, scaling much better to larger trees and datasets than the method proposed in \cite{irsoy-soft-2012}. Our method for growing also incorporates regularization terms that were not considered previously. While we also use first-order methods to train our models, there have been significant advances in first-order optimization methods since the original soft trees paper was published, considerably improving the efficiency and quality of the solutions found \cite{loshchilov-decoupled-2019, kingma-adam-2017}. We also optimize over the entire tree structure, as well as optimizing over each new split, which was not considered in the original work. As a result, our method yields models that are often competitive with or outperform tree ensemble methods, whereas the original work did not compare against these ensemble methods, and was only shown to be competitive with linear discriminant trees.

A more modern treatment of soft decision trees is provided in \cite{frosst_distilling_2017}, which considers various new aspects such as regularization and knowledge distillation from neural networks. However, this work does not consider growing the trees and only considers hyperplane split functions. They also treat the leaf nodes as ``bigots'', namely leaf nodes that represent determined classes or probability distributions. By learning linear functions in the leaf nodes, we can show that learning ``bigots'' is just a special case of our model formulation. We also introduce another regularization term, the sample penalty, that improves both accuracy and interpretability. Importantly, the results in \cite{frosst_distilling_2017} rely on knowledge distillation from neural networks. We do not use any form of knowledge distillation and achieve higher accuracy on the MNIST dataset than this work reports.

There are also papers that predate \cite{irsoy-soft-2012}, such as \cite{olaru_complete_2003}, which introduce some ideas of growing ``fuzzy'' decision trees, but they do not consider anything beyond parallel splits, and they do not use gradient descent. The idea of using neural networks as split functions was first introduced by \cite{guo_classification_1992}, which uses shallow multi-layer perceptrons to learn non-linear split functions in the nodes. However, this work considered hard-split decision trees, not soft trees, and did not investigate other non-linear split functions, such as convolutional layers, which we explore in this work. The idea of incorporating elements from deep learning into tree structures was also inspired by the theoretical equivalence in modeling power between neural networks and Optimal Classification Trees with Hyperplanes (\cite{bertsimas2019machine}). Another approach of introducing nonlinearity in the tree nodes was introduced by \cite{bertsimas_near-optimal_2021}. This work focuses on Regression Trees and employs hyperplane splits for the inner nodes and a polynomial prediction function for the leaf nodes. Our approach introduces nonlinearity in the inner nodes instead of the leaves, thus further increasing modeling capacity. As a tree-based method, our work is also related to ensemble tree methods, such as Random Forests and Boosted trees \cite{random-forest, xgboost}. We compare our approach against these ensemble methods and also demonstrate how our approach is significantly more interpretable.

\section{Tractability: Methods of Generalized Soft Trees}\label{sec:methods}

In this section, we present a class of differentiable soft decision tree models that can be trained via backpropagation and gradient descent methods. Critically, we consider a broad class of differentiable split functions that go beyond traditional parallel and hyperplane splits. Further, we develop a method for \textit{growing} these trees, allowing us to create unbalanced tree structures that are potentially more suited for a given task or dataset.

\subsection{Generalized Soft Trees}\label{sec:gst}
 We now introduce our model class, the Generalized Soft Tree (GST). Structurally, we consider a tree $\mathcal{T}$, and its set of leaves $\mathcal{L}$. Each leaf node $\ell \in \mathcal{L}$ can be reached through a path $P_{\ell}$, which starts from the root node and ends at the leaf node $\ell$. We consider $P_\ell$ as just the set of internal nodes that lead to leaf $\ell$ (we exclude the actual leaf node).

Each node $n \in P_{\ell}$ is associated with a function $f_{n}(\boldsymbol{x}; \theta_n):  \mathbb{R}^n \rightarrow \mathbb{R}$. If $f_n(\boldsymbol{x}; \theta_n) \leq 0$, the next node of the path is the left child of node $n$, otherwise it is the right child, like in typical decision trees. Thus, to reach a specific leaf $\ell$, there are some nodes in $P_\ell$ that are then followed by their left child (set of nodes $L_\ell$) and the rest by their right child (set of nodes $R_\ell$). Each leaf $\ell$ is also associated with a function $g_{\ell}(\boldsymbol{x}; \pi_\ell):  \mathbb{R}^n \rightarrow \mathbb{R}^d$. In our work, we consider $g_\ell$ to be a linear function $\boldsymbol{a}_\ell^{\intercal} \boldsymbol{x} + b_{\ell}$, where $\boldsymbol{a}_\ell$ and $b_\ell$ are trainable.

Then, a sample of our dataset $\boldsymbol{x}$ ends up in leaf $\ell$ if and only if the product $A(\boldsymbol{x}) = \prod_{n1 \in L_\ell} \mathbb{1} \{f_n(\boldsymbol{x}) \leq 0 \} \cdot \prod_{n \in R_\ell} \mathbb{1} \{f_n(\boldsymbol{x}) > 0 \} = 1$.
We first define an auxiliary function $\Phi(\boldsymbol{x})$:

\begin{equation}
    \Phi(\boldsymbol{x}) = \sum_{\ell \in \mathcal{L}} \bigg ( (\boldsymbol{a}_{\ell}^{\intercal} \boldsymbol{x} + b_{\ell}) \cdot \prod_{n \in L_\ell} \mathbb{1} \{f_n(\boldsymbol{x}) \leq 0 \} \cdot \prod_{n \in R_\ell} \mathbb{1} \{f_n(\boldsymbol{x}) > 0 \}  \bigg ).
\end{equation}

If the sample belongs to leaf $\ell^*$, then the value of the function $\Phi(\boldsymbol{x})$ is equal to $\boldsymbol{a}_{\ell^*}^{\intercal} \boldsymbol{x} + b_{{\ell^*}}$. This is hence the output of our model.

Using this output $\Phi(\boldsymbol{x})$, we employ the MSE loss function for regression problems and the Cross-Entropy loss function for classification problems. However, in order to be able to utilize the backpropagation algorithm to optimize the model, we need $\Phi(\boldsymbol{x})$ to be differentiable. We therefore modify slightly our leaves' assignment function $A(\boldsymbol{x})$ to a softer version, where we replace the non-differentiable indicator function with the sigmoid function:

\begin{equation}
    \sigma(\boldsymbol{x}) = \frac{1}{1+ \exp^{-\boldsymbol{x}}}.
\end{equation}

So, our final version of $\Phi(\boldsymbol{x})$ is:

\begin{equation}
    \Phi(\boldsymbol{x}) = \sum_{\ell \in \mathcal{L}} \bigg( (\boldsymbol{a}_{\ell}^{\intercal} \boldsymbol{x} + b_{\ell}) \cdot \prod_{n \in L_\ell}(1 - \sigma (f_n(\boldsymbol{x}))) \cdot \prod_{n \in R_\ell} \sigma (f_n(\boldsymbol{x}))  \bigg ).
\end{equation}

Trees can be either full (balanced) or unbalanced. In general, we create full trees and then grow them in an unbalanced way, as described in Section \ref{deeptree-algo}. Regardless of the structure of the actual tree, the input $\boldsymbol{x}$ is passed through and processed at every node of the tree, and each calculation to determine the split is independent from the rest of the tree. For this reason, we adopt a parallelized approach on these calculations, which we observe critically reduces the training time of the model. 



\subsection{Regularization}\label{subsec:regularization}

Simply employing the appropriate loss function (MSE or Cross Entropy) does not give us any control on how the samples are distributed to the leaves. We observe that in many cases, samples tend to concentrate in a small subset of the leaves, which means that a large part of the tree is underutilized. To address this issue, we employ regularization, and more specifically we add a term to ensure that samples are distributed to the leaves in a more balanced way. This approach, however, softens the class assignments, since each sample tends to be assigned simultaneously and partially to multiple leaves. We therefore need to limit the number of leaves it essentially belongs to and achieve this goal with the two regularization techniques described below. 

We first start with a useful definition of leaf weights. If we have $N$ samples $\boldsymbol{x_i}$, $i=1,\dots,N$, we define the leaf weight $w_{i,\ell}$ as:

\begin{equation}
    w_{i,\ell} = \prod_{n \in L_l}(1 - \sigma (f_n(\boldsymbol{x}_i))) \cdot \prod_{n \in R_l} \sigma (f_n(\boldsymbol{x}_i)).
\end{equation}

This quantifies how much sample $i$ is assigned to leaf $\ell$, since we have a soft tree. Thus, for leaf $\ell$, the sum $S_\ell = \sum_{i=1}^{n} w_{i,\ell}$ is analogous to the number of samples that fall into leaf $\ell$ in a standard decision tree.


As discussed in Section \ref{deeptree-algo}, we can use $S_l$ to parameterize our tree-growing algorithm similar to the minimum bucket parameter in standard CART algorithms \cite{cart}. 
\begin{itemize}
    \item \textbf{(Split Penalty) Uniform sample distribution over leaves}. To avoid overfitting and getting trapped in local optima, we want to encourage the tree to use all of the leaves with the number of samples per leaf $S_\ell$, $\ell \in \mathcal{L}$ distributed evenly. To achieve this, we normalize the sum of the leaf weights, resulting in a distribution over the leaf nodes:
    $$\boldsymbol{P} = \bigg \{ \frac{S_\ell}{\sum_\ell S_\ell}\bigg \}_{\ell \in \mathcal{L}}.$$
    Our penalty becomes the cross-entropy loss between $\boldsymbol{P}$ and the discrete uniform distribution $\boldsymbol{U} = \big \{ \frac{1}{|\mathcal{L}|}\big \}_{\ell \in \mathcal{L}}$, defined as
    \begin{equation}
        L_{\boldsymbol{U}} = - \sum_{\ell \in \mathcal{L}}\frac{1}{|\mathcal{L}|}\log(P_\ell) + \left( 1 - \frac{1}{|\mathcal{L}|} \right) \log(1 - P_\ell).
    \end{equation}

    Alternatively, we experiment with using squared difference instead of cross entropy:

    \begin{equation}
        L_{\boldsymbol{U}} = \sum_{\ell \in \mathcal{L}} \left( P_\ell - \frac{1}{|\mathcal{L}|} \right) ^2.
    \end{equation}

    \item \textbf{(Sample Penalty) Sparsity in leaf distribution per sample}. While we want to encourage each leaf to contribute to model predictions, we also want to minimize the number of leaves used to make a prediction for a given sample. This helps preserve interpretability and encourages the model to learn meaningful splits, rather than spreading the prediction for each sample across all the leaf nodes evenly. To achieve this, for a given sample $i$, we use the weight distribution over leaf nodes,
    $$\boldsymbol{Q_i} = \big \{w_{i,\ell}\big \}_{\ell \in \mathcal{L}}.$$
    We want to encourage $\boldsymbol{Q_i}$ to be as far from the uniform distribution as possible, so we define our penalty as the cross entropy between  $\boldsymbol{Q_i}$ and the uniform distribution $\boldsymbol{U} = \big \{ \frac{1}{|\mathcal{L}|}\big \}_{l \in \mathcal{L}}$ as follows:
    \begin{equation}
        L_S =  \frac{1}{N}\sum_{i=1}^N\sum_{\ell \in \mathcal{L}}\frac{1}{|\mathcal{\mathcal{L}}|}\log(Q_{i,\ell}) + \left(1 - \frac{1}{|\mathcal{L}|} \right)\log(1 - Q_{i,\ell}).
    \end{equation}
\end{itemize}
With these penalty terms defined, we can then give the overall loss function as

\begin{equation}
    L(X,y,T) = \ell(T(X), y) + \alpha_UL_{\boldsymbol{U}}(T,X) + \alpha_SL_S(T,X),
\end{equation}
where $\alpha_U$ and $\alpha_S$ are parameters to adjust the magnitude of the penalties. We will discuss these parameters further in Section \ref{deeptree-algo}, where we treat these parameters as dynamic and update them as a function of the training step and tree depth. 

\subsection{Hyperplane Trees}

In this version of GSTs, the functions of the inner nodes are linear functions with trainable parameters. More specifically, we define $f_n(\boldsymbol{x}; \theta_n)$ as:

\begin{equation}
    f_n(\boldsymbol{x}; \boldsymbol{a}_n, b_{n}) = \boldsymbol{a}_{n}^{\intercal} \boldsymbol{x} + b_{n}.
\end{equation}

So, the split is determined by the sign of $f_n(\boldsymbol{x}; \boldsymbol{a}_n, b_{n})$. In this case, the weight $a_{nj}$ that multiplies each feature $x_j$ specifies how important this feature is for the specific split. We refer to nodes that split on such functions as hyperplane nodes.

\subsection{Convolutional Trees} \label{subsec:conv trees}

In this variation of a GST, each inner node (all non-leaf nodes) is essentially a layer of a Convolutional Neural Network. It consists of a convolutional layer, a ReLU activation layer, a max pooling operation, and finally a linear layer, which transforms the resulting map to a positive/negative number, which is then used for the split. We refer to nodes with these convolutional operations as convolutional nodes. Each leaf node is a hyperplane node, since we observe that making leaf nodes convolutional does not improve the performance of the tree. 

We now further explain each operation that takes place inside inner nodes. 

\begin{itemize}
    \item \textbf{Convolutional layer}. This layer performs a spatial (2D) convolution operation between the input image and a kernel, which is a set of weights of fixed dimension that slides across the image. Each node of the tree is associated with a kernel with different trainable weights. In essence, each kernel learns a different pattern inside the image. The parameters of this operation are:
    \begin{itemize}
        \item The kernel size of node $n$, namely the dimensions $W_n$ (width), $H_n$ (height) and $D_n = D$, which is equal to the depth of the image,
        \item The stride $S$, which refers to the number of spatial incremental steps that are used when the kernel is slided vertically and horizontally across the image, and
        \item The padding $P$, which refers to the rows and columns of zeros that can be added to the original image, in order to preserve its shape after the convolution. 
    \end{itemize}

    If $I_{x,y,d}$ denotes the image value at position $(x,y)$ and depth $d$, $d=1,\dots, D$, and $v_{n,d}$, $d=1,\dots, D$, denotes the weights of the kernel in the node $n$, arranged in the shape $W_n \times H_n$, then the convolution result $C_{n, x, y}$ of kernel $n$ with the input image at $(x,y)$ is calculated as:

    \begin{equation}
        C_{n, x, y} = \sum_{d=1}^{D}v_{n,d}*I_{x,y,d} =\sum_{d=1}^{D} \sum_{i = -(W_n-1)/2}^{(W_n-1)/2} \sum_{j = -(H_n-1)/2}^{(H_n-1)/2}v_{n,i,j,d} \cdot I_{x-i, y-j, d},
    \end{equation}
    
    where $i$ and $j$ span the dimensions of the kernel. If we add a bias $u_{n,d}$ to the result above, we get $z_{n,x,y}$:
    
    \begin{equation}
        z_{n,x,y} = \sum_{d=1}^{D}[v_{n,d}*I_{x,y,d} + u_{n,d}] = C_{n,x,y} +  \sum_{d=1}^{D} u_{n,d}.
    \end{equation}

    \item \textbf{Activation layer}. This layer feeds the convolution output $z_n$ to a ReLU activation function and adds nonlinearity to increase modeling power:

    \begin{equation}
        o_{n,x,y} = \max(z_{n,x,y},0).
    \end{equation}

    \item \textbf{Pooling layer}. This operation subsamples the output $o_n$ to achieve translational invariance, meaning that if a small translation occurs, most of the pooled values won’t change. This property is important, especially if we are interested in whether a specific feature is present in the image, rather than where exactly it is. In our case, we employ max pooling, where the maximum value of a $2 \times 2$ neighborhood is selected, using a stride of 2:

    \begin{equation}
        P_{x,y,d} = \max_{i,j=0,\dots,1}o_{2x+i, 2y+j, d}.
    \end{equation}

    This operation reduces the size of each dimension by half. 
    
    \item \textbf{Linear layer}. After the sequence of the operations already mentioned, the weighted sum of the output is calculated and the split is determined by its sign:

    \begin{equation}
         , 
    \end{equation}
     where $\boldsymbol{P}$ is flattened.

     If we want to omit the last linear layer, we can just calculate the sum of the values of $\boldsymbol{P}$ and add a bias:

     \begin{equation}
        f_n(I;v_n, u_n, b_n) = \boldsymbol{e}^\intercal \boldsymbol{P} + b_n.
    \end{equation}
\end{itemize}

\section{Tractability: Efficiently Growing Soft Trees}\label{deeptree-algo}
Often, previous work considers soft decision trees only as fixed, full trees \cite{hinton_distilling_2015}. Constraining the structure of a tree and the way it is trained, however, can be significantly limiting. A full tree of depth $d$ has $2^d$ leaves, so we are limited to a collection of tree structures where the number of leaves grows exponentially with depth. This increasing gap in the size of trees as we increase depth makes it very likely that the optimal number of leaves and tree structure falls somewhere between two tree structures (of depth $d$ and $d+1$). We will later show empirically that such a situation does indeed occur in real-world datasets. It is also not certain that at a fixed depth $d$, a full tree is an optimal structure. 

Other works on soft trees, including the original paper on the topic, develop a method for growing soft trees starting from one root node \cite{irsoy-soft-2012}. The soft trees are then grown recursively, creating and training a new split at each leaf node until the validation loss no longer decreases. While this approach may work for small trees, simple split functions, and small feature spaces, exhaustively searching over all leaf nodes and training a new split for each becomes intractable as the tree becomes larger and the number of parameters in the split functions increases. To address these limitations, we develop a heuristic method for growing soft decision trees which we will refer to as the DeepTree algorithm.
\subsection{The DeepTree Algorithm}
In this section, we develop our approach to growing soft decision trees, which we call the DeepTree algorithm. For each step of our algorithm, the core of our approach can be generally outlined as follows:
\begin{itemize}
    \item \textbf{Identify the best leaf to split on}. At each step, we select some leaf node that we are going to replace with a new split. Importantly, rather than iterating over all leaf nodes, training new splits, and evaluating whether they are worth keeping, we simply select the leaf that has the highest potential for loss reduction.  
    \item \textbf{Create and Train a new subtree}. Once we have selected where the new split will occur, we create a new depth-$1$ subtree. We also train this new subtree before adding it to the main tree to provide a warm-start.
    \item \textbf{Insert the new subtree}. Update the main tree by replacing the selected leaf node with the new depth-$1$ subtree.
\end{itemize}
Each of these steps requires careful consideration, and we will build up to our algorithm by walking through our approach to each of them. The first task of identifying the best leaf on which to split is arguably the most delicate and subtle of the steps above. To begin, we need a measure of the loss in each leaf $\ell$, which we define as 
\begin{equation}
    L_\ell(X, y) = \sum_{j=1}^N w_{\ell,j}\ell(g_\ell(x_j; \pi_\ell), y_j),
\end{equation}
which is the soft-tree analog of the leaf loss in a standard decision tree. Note that we do not normalize the leaf loss by the total weight of the samples in the leaf. Computing this leaf loss across the leaves is not a challenge, so we can readily calculate and access them at each step in our algorithm. However, if we were to try to follow the rest of the procedure for growing a standard decision tree, we quickly run into major computational challenges. Ideally, we would evaluate each possible new split and select the one that minimizes the loss. Creating and training a new split requires creating and training a new depth-$1$ subtree containing an inner node and two leaf nodes, each with at least $n$ parameters, which are trained jointly. While this training process can be optimized significantly, the run-time to train a single depth-$1$ tree can still be on the order of O(1) seconds even when trained on a dedicated GPU. Therefore, if we were to adopt this procedure for selecting each node to split on, we would have to train potentially hundreds or thousands of depth-$1$ subtrees at each iteration, which could take O-10 min. While technically feasible, we believe this is an impractical approach to growing soft decision trees. Instead, we adopt the simple heuristic of selecting the leaf with the largest leaf-loss,
\begin{equation}
    \ell_{split} = \text{argmax}_{l : S_\ell \geq w_{\min}} \; L_\ell(X,y),
\end{equation}
restricting ourselves to the set $\{\ell : S_\ell \geq w_{\min}\}$, where $w_{\min}$ is a minimum leaf weight to be eligible for splitting, which is a hyperparameter that can be tuned by the user. While this approach to selecting the leaf to split on may not be optimal, it also carries far less risk of over-fitting compared to choosing a bad leaf to split on in a standard decision tree. In the growing of a standard parallel-split decision tree, creating a new split necessarily involves partitioning the data in a leaf, and making predictions in the new leaves based on this partition. This can quickly lead to over-fitting if a meaningful partition (one that captures some signal) does not exist. In contrast, for our GST models, a leaf takes on the parameterized form $g_{\ell}(\boldsymbol{x}; \pi_\ell) = \boldsymbol{a}_\ell^{\intercal} \boldsymbol{x} + b_{\ell}$. Then for any split comprised of a new inner node function $f_s$ and two leaf nodes with functions $g_{s_1}$ and $g_{s_2}$, regardless of what is learned for $f_s$, it is possible to learn $g_{s_1} = g_{s_2} = g_l$. Therefore, a split can be learned which is equivalent to the original leaf node. Assuming sufficient regularization and training with early stopping, tracking the loss on some validation set, it is unlikely that our approach to selecting split locations will lead to over-fitting. 

\hfill

The next step in our approach is creating and training the new subtree that will be added to our current tree. We refer to the tree in its current state as $\mathcal{T}$ and the new subtree as $\mathcal{T}_s$. We instantiate $\mathcal{T}_s$ as a new depth-$1$ subtree. We want to train $\mathcal{T}_s$ as if it were integrated into $\mathcal{T}$, where $\mathcal{T}$ is fixed. To do this, we first compute the leaf weight $w_{s, j}$ of the split leaf $l_s$ for each input sample $x_j \in X$. Letting $f_s$ represent the inner node for $\mathcal{T}_s$, and $g_{s_1}$ and $g_{s_2}$ represent the new leaf nodes for $\mathcal{T}_s$, we compute the weighted output for the new subtree as:
\begin{equation}
    \mathcal{T}_s(x_j) = w_{s, j} (\sigma (f_k(x_j ; \theta_{s_1}))g_{s_1}(x_j; \pi_{s_1}) + (1 - \sigma (f_k(x_j ; \theta_{s_2})))g_{s_2}(x_j; \pi_{s_2})).
\end{equation}
Let us denote the output of the current tree $\mathcal{T}$ without the contribution of the leaf that is being split on by $\mathcal{T}_{\neg s}$, so that:
\begin{equation}
        \mathcal{T}_{\neg s}(X) \stackrel{\text{def}}{=} \sum_{l \in \mathcal{L} \setminus l_s} \bigg( (\boldsymbol{a}_{l}^{\intercal} \boldsymbol{x} + b_{l}) \cdot \prod_{n \in L_l}(1 - \sigma (f_n(\boldsymbol{x}))) \cdot \prod_{n \in R_l} \sigma (f_n(\boldsymbol{x}))  \bigg ).
\end{equation}
We then train the subtree as we would normally, now computing the loss as:
\begin{equation}
    L(X, y, \mathcal{T}_s) =L(\mathcal{T}_s(X) + \mathcal{T}_{\neg s}(X), y) + \alpha_U L_U (\mathcal{T}_s, X) + \alpha_S L_S (\mathcal{T}_s, X),
\end{equation}
where $\mathcal{T}(X)$ is the fixed output of the current tree $\mathcal{T}$. Training $\mathcal{T}_s$ in this way is equivalent to merging $\mathcal{T}_s$ into $\mathcal{T}$, freezing the parameters of $\mathcal{T}_{\neg s}$, and training only the parameters in $\mathcal{T}_s$. In our experiments, it was more convenient to train the new subtree in this way and then merge, rather than merge and then train, but both approaches are technically feasible.

\hfill

What remains is merging the new subtree $\mathcal{T}_s$ into the existing tree $\mathcal{T}$. However, this is purely an engineering matter, and depends on how the tree is implemented. In any case, this operation can be done efficiently by merging the weights from the two trees, which can be done in linear time in the number of weights in the worst-case. Combining these steps, we arrive at the DeepTree Algorithm, given below.

\begin{algorithm}[H]
\begin{algorithmic}[1]
\STATE \textbf{Input Data} ($\mathcal{T}$, $X$, $y$, max\_iters, retrain\_steps, $w_{\min}$, $\alpha_U, \alpha_S$, $L$)

\STATE it $\leftarrow 0$

\STATE $W \leftarrow \text{GetLeafWeights}(\mathcal{T}, X)$
\STATE $\ell_{split}$ $\leftarrow$ BestSplitLeaf($X$, $y$, $W$, $w_{\min}$)

\WHILE{it $<$ max\_iters \AND $\ell_{split}$ exists}
\STATE $\text{it} \leftarrow \text{it}+1$
\STATE $\mathcal{T}_s \leftarrow \text{CreateTrainSubtree}(X, y, W, \mathcal{T}, \alpha_U, \alpha_S, L)$
\STATE $\mathcal{T} \leftarrow \text{InsertSubtree}(\mathcal{T}, \mathcal{T}_s)$
\IF{it \% retrain\_steps == 0}
\STATE $\mathcal{T} \leftarrow$ train($\mathcal{T}$, X, y, $\alpha_U, \alpha_S$, $L$)
\ENDIF
\STATE $W \leftarrow \text{GetLeafWeights}(\mathcal{T}, X)$
\STATE $\ell_{split} \leftarrow$ BestSplitLeaf($X$, $y$, $W$, $w_{\min}$)

\ENDWHILE
\end{algorithmic}
\caption{DeepTree Algorithm}
\label{alg:deeptree}
\end{algorithm}

\begin{algorithm}[H]
\begin{algorithmic}[1]
\STATE \textbf{Input Data} ($X$, $y$, $\mathcal{T}$, $W$, $w_{\min}$)
\STATE $\mathcal{S} \leftarrow \{\}$
\FOR{Leaf node $\ell$ in  $T$}
\STATE $S_\ell \leftarrow \sum_{i=1}^n W_{i,\ell}$
\IF{$S_\ell \geq w_{\min}$}
\STATE $\mathcal{S} \leftarrow \mathcal{S} \cup \{\ell\}$
\ENDIF
\ENDFOR
\STATE $\ell_{split} \leftarrow \text{argmax}_{\ell \in S} L_\ell(X,y; W)$
\RETURN $\ell_{split}$

\end{algorithmic}
\caption{BestSplitLeaf}
\label{alg:splitleaf}
\end{algorithm}

\begin{algorithm}[H]
\begin{algorithmic}[1]
\STATE \textbf{Input Data} ($X, y, W, \mathcal{T}, \alpha_U, \alpha_S$, $L$, $\ell_{split}$)
\STATE $\mathcal{T}_s \leftarrow \text{New depth-$1$ tree}$
\STATE $\tilde y \leftarrow \mathcal{T}_{\neg s}(X)$
\STATE $L_s \leftarrow L_s(\; \cdot \; ; L, \tilde y)$
\STATE $\mathcal{T}_s \leftarrow \text{train}(\mathcal{T}_s, X, y, \alpha_U, \alpha_S, L_s)$
\RETURN $\mathcal{T}_s$
\end{algorithmic}
\caption{CreateTrainSubtree}
\label{alg:train-subtree}
\end{algorithm}

We do not define the GetLeafWeights algorithm as it is simply extracting the leaf weights $W = [w_{i,\ell}]$. We also reference a \textbf{train} function, which we do not define in pseudo-code as this function depends on the implementation, and is simply a training loop. In practice, the \textbf{train} function will also include a variety of hyperparameters, such as the learning rate and number of epochs, which we omit in Algorithm \ref{alg:deeptree} for brevity. We also introduce parameters max\_iters and retrain\_steps, which control the total number of leaf nodes to add, and how frequently we should refit the whole model, respectively. In the following section, we present the results of training full trees and applying this method for growing soft trees to a variety of classification tasks with different data modalities. 

\section{Performance: Experimental Datasets}\label{sec:results}
In this section, we evaluate our GST models on a variety of tabular and image datasets, demonstrating their ability to handle both structured and unstructured data. Further, we evaluate our DeepTree algorithm as a method for learning more efficient and accurate GSTs. We compare our results against the state-of-the-art for each dataset, as well as other existing tree-based methods.

\subsection{Training methodology}\label{subsec:training-methodology}

Before presenting the specific results, we will outline the experimental set-up and the purpose of the different experiments we ran. For each dataset, we use a training set, used to train the tree, and a test set, to evaluate the trained model out-of-sample. The training process involves selecting a set of hyperparameters, which are dataset-specific. In order to tune most of these hyperparameters, we further split the training set into a training and a validation set, where the validation set is not used for the training phase but to monitor the performance of the model out-of-sample during training. Typically, 10\% of the training set is selected as validation set.

\begin{itemize}
    \item \textbf{Depth of the tree.} This is the most important hyperparameter, since it determines the size of our model. For each dataset, we experimented with 7 different depth values (4-10), selected based on the dataset size and the number of features, to explore how the size of the models impacts performance. 
    \item \textbf{Learning rate.} This parameter controls how large the step of the optimization algorithm we use for backpropagation is. For each dataset, we want to find the appropriate value for the learning rate, so that it is neither too big, which results in the algorithm being stuck in local minima, nor too small, which means that the convergence of the algorithm is unnecessarily slow. 
    \item \textbf{Batch size.} The batch size determines how many samples are used in parallel through the forward pass of the model. The backpropagation is also applied per batch, so the model parameters are updated after each batch of training data is passed through the tree. It is selected based on the dataset size and the memory capacity; in general, a very small batch size slows down the training process. 
    \item \textbf{Weight decay.} This parameter scales an $L_2$-regularization term of the weights that are used in each node. This term is added to the loss function to make the training process more stable and prevent the weights from taking too large values.
    \item \textbf{Scaling factor $\alpha_U$.} This is a hyperparameter we introduce to weigh the regularization term associated with the uniform sample distribution over the leaves added in the loss function, as described in Section \ref{subsec:regularization}. 
    \item \textbf{Scaling factor $\alpha_S$}. This hyperparameter is used to weigh the regularization term that forces sparsity in the leaf distribution for each sample. Its purpose is also discussed further in Section \ref{subsec:regularization}.
    \item \textbf{Dropout rate.} Dropout is a technique used to prevent overfitting of the model on the training set, by randomly ignoring some of the node weights at each training epoch. The dropout rate determines the probability that each node weight is zeroed.
    \item \textbf{Number of epochs.} The number of epochs is a very dataset-specific parameter. It is hard to tune, since if the model is trained for too many epochs, overfitting on the training set occurs, but also if we select a conservative number, the model might not fully capture the underlying structure of the training data. For this reason, we employ early stopping, which uses the validation set to terminate the training process when a fluctuation in the validation loss is observed during training. This fluctuation indicates that further training the model does not actually improve its out-of-sample generalization ability.
\end{itemize}

In order to tune the aforementioned hyperparameters, we use an automated search procedure which employs Bayesian Optimization \cite{bayesian-opt} to explore the parameter space, thus avoiding the generally more time-consuming and limiting Grid Search approach. For this procedure, we specify a range for each hyperparameter, as well as the number of the random runs until the best hyperparameters are finalized. For each run, the hyperparameters are selected based on the previous runs, trying to optimize over the validation loss. To make this process more robust, we perform cross-validation and we aim to optimize the average validation loss that is obtained across the different folds. 

After finding the best hyperparameters, we train full trees at 7 different depths, to explore how increasing the size of the tree affects its performance. We present an average accuracy/AUC of the trees, to ensure that our results are more robust. 

Finally, we demonstrate our technique of first growing a full tree and then moving to a grown, unbalanced tree with larger depth. We show how this procedure improves performance by growing each aforementioned full tree and compare the resulting grown tree to the original full tree, as well as full trees of bigger depth. We add leaves to full trees according to Table \ref{tab:num_of_leaves}. Naturally, as original depth of the tree increases, we add more leaves; we observe that adding a few leaves is sufficient, even when the original depth is large.   

From all the models we have obtained through the experiments, we select the best (again based on the validation set) and compare its test set performance to other tree-like methods, namely typical Decision Trees, Random Forests and XGBoost. 

\begin{table}[!ht]\setlength\extrarowheight{1.7pt}
\centering
\begin{tabular}{|c|c|}
\hline
\textbf{Original Depth} & \textbf{Number of leaves added} \\
\hline
4 & 8 \\\hline
5, 6 & 16 \\ \hline
7-10  &32 \\ 
\hline
\end{tabular}
\caption{Number of leaves added for each depth of an original full tree.}\label{tab:num_of_leaves}%
\end{table}

\subsection{Tabular Data}\label{subsec:tab}
\textbf{MIMIC IV.} The MIMIC IV dataset contains critical care data for more than 40,000 patients admitted to intensive care units (ICU) at the Beth Israel Deaconess Medical Center. We select 70,887 patients who were admitted to the ICU and predict their mortality risk using 14 risk factors including age, vitals, and blood panel data. We compare the performance of general tree methods to full and grown GSTs in mortality classification. For GST models, five trees were trained for each of the tree depths using the approach outlined in Section \ref{subsec:training-methodology}, and the average out-of-sample AUC scores are presented in Table \ref{tab:mimic_full_tree}. The results show an increase in out-of-sample AUC when growing the trees deeper. Additionally, we observe that among GSTs of the same depth, GSTs trained with sample penalty consistently outperform those trained without sample penalty.
\begin{table}[!ht]\setlength\extrarowheight{1.7pt}
\centering
\begin{tabular}{|c|c|c|c|c|}
\hline
\textbf{Depth} & 
\begin{tabular}{@{}c@{}} \textbf{Average AUC}  \\\textbf{(\%)}  \end{tabular} &
\begin{tabular}{@{}c@{}} \textbf{Average AUC, } \\ \textbf{sample penalty (\%)}  \end{tabular} \\
\hline
 $4$ & $68.91 (\pm 0.22)$ & $70.76 (\pm 0.11)$  \\
 $6$ & $71.01 (\pm 0.12)$  &$71.42(\pm 0.30)$ \\ 
 $8$ & $71.61 (\pm 0.81)$  &$73.28 (\pm 0.54)$ \\ 
 $10$ & $\mathbf{72.73 (\pm 0.27)}$  &$ \mathbf{73.45 (\pm 0.49)}$ \\ 
\hline
\end{tabular}
\caption{MIMIC-IV: Average out-of-sample AUC scores of full GSTs trained with and without sample penalty. AUC increases as depth increases and when trained with sample penalty.}\label{tab:mimic_full_tree}%
\end{table}
\begin{table}[!ht]\setlength\extrarowheight{1.7pt}
\centering
\begin{tabular}{|c|c|c|c|}
\hline
Depth &Additional leaves&   Average AUC (\%) \\
\hline  
4 &8 &$72.58(\pm 0.24)$ \\ 
6  &16&$72.47(\pm 0.36)$ \\
8 &32& $73.87(\pm 0.22)$   \\ 
10 &32& $\mathbf{73.96(\pm 0.10)}$  \\ \hline
\end{tabular} 
\caption{MIMIC-IV: Average out-of-sample AUC scores of grown GSTs models.}\label{tab:mimic_grown_tree}%
\end{table}
We grow each of the full GSTs trained with sample penalty using Algorithm \ref{alg:deeptree} and compare the average AUC in Table \ref{tab:mimic_grown_tree}. We can observe that adding leaves using this approach generally improves the AUC of the models. It is also worth noting that adding a small number of leaves to a shallow tree using our approach can achieve similar level of AUC of deeper full GSTs. For example, adding 8 leaves to a tree of depth 4 can outperform some trees of depth 8. 

\begin{table}[!ht]\setlength\extrarowheight{1.7pt}
\centering
\begin{tabular}{|c|c|c|c|}
\hline
Model & Best model AUC (\%) \\
\hline
Hyperplane GST & $74.00$\\ \hline
CART &$67.43$ \\ \hline
Random Forest  &$74.70$ \\ \hline
XGBoost  & $\mathbf{75.06}$   \\ 
\hline
\end{tabular}
\caption{MIMIC-IV: Best out-of-sample AUC scores of general tree methods.}\label{tab:mimic_other_tree}%
\end{table}
In addition to our GST models, we compare the performance of general tree models (CART, Random Forest, and XGBoost) on this dataset. For each model, we performed a hyperparameter search with 3-fold cross-validation, and report results using the best parameters selected for each model in Table \ref{tab:mimic_other_tree}. We observe that GSTs have comparable performance to Random Forest.

\par\noindent \\
\textbf{MNIST.} MNIST is one of the most popular datasets in machine learning, consisting of grayscale images that contain handwritten digits (0-9) \cite{mnist-original}. The simplicity of these images and their relatively small resolution (28x28) have resulted in MNIST often being handled as a tabular dataset, where each sample is a vector resulting from flattening the images. In our experiments, Hyperplane GSTs are very successful, but we also consider Convolutional GSTs in order to compare performance.

MNIST consists of 60,000 training images and 10,000 test images and is balanced across the different classes. The state-of-the-art out-of-sample accuracy is 99.87\%, using CNNs and Homogeneous Vector Capsules \cite{mnist-bench}. 

We first experiment with depth-4 full trees, in order to get the best hyperparameters for the training process of the trees. We use the hyperparameter tuning method presented in Section \ref{subsec:training-methodology} and we specify the best hyperparameters using a 3-fold cross-validation approach to calculate the average validation loss through 100 random runs. With these hyperparameters, we then train trees with depth ranging from 4 to 10. In order to ensure the stability of reported results, we train 5 trees for each depth. For training the models, we split the training set to training and validation set, which is used to monitor the generalization ability of the model, and to determine the early stopping of the training process. We report the average out-of-sample accuracy in Table \ref{tab:MNIST} for selected depths. We select accuracy as our metric since we have a quite balanced multiclassification problem.

\begin{table}[!ht]\setlength\extrarowheight{1.7pt}
\centering
\begin{tabular}{|c|c|c|c|c|c|}
\hline
Type & Depth & 
\begin{tabular}{@{}c@{}} Average Accuracy \\ (Full) (\%)  \end{tabular} &
\begin{tabular}{@{}c@{}} Average Accuracy \\ (Grown) (\%)  \end{tabular} \\
\hline
Hyperplane & $4$ & $97.64 (\pm 0.10)$  & $97.70 (\pm 0.09)$ \\
& $6$     & $98.15 (\pm 0.11)$  & $98.14 (\pm 0.09)$   \\
& $8$    & $98.27 (\pm 0.09)$  & $98.25 (\pm 0.07)$ \\
& $10$   & $\mathbf{98.34 (\pm 0.18)}$  & $\mathbf{98.36 (\pm 0.13)}$\\ \hline

Convolutional & $4$ & $98.16 (\pm 0.06)$  & $98.15 (\pm 0.07)$  \\
& $6$     & $98.43 (\pm 0.09)$  & $98.49 (\pm 0.03)$ \\
& $8$    & $98.74 (\pm 0.06)$  & $98.72 (\pm 0.08)$ \\
& $10$   & $\mathbf{98.86 (\pm 0.03)}$ & $\mathbf{98.87 (\pm 0.04)}$ \\ \hline

\end{tabular}
\caption{MNIST Full trees accuracy (Hyperplane and Convolutional Trees). Average accuracy and standard deviation are calculated across 5 different trees.}\label{tab:MNIST}%
\end{table}

For Hyperplane GSTs, we observe that by increasing the depth of the tree, the accuracy increases monotonically. It is remarkable that our results are comparable to the state-of-the-art, without leveraging any image processing or deep learning techniques, and by using considerably less parameters. Convolutional GSTs further improve upon Hyperplane GSTs, and they also yield better results as their depth is growing, but their improvement margin is small, indicating that Hyperplance GSTs are powerful enough to handle the dataset's complexity. 

In order to see if further growing the resulting full trees improves performance, we use the 5 trees we train for each depth and we grow them, using the method described in Section \ref{deeptree-algo} and adding the number of leaves described in Table \ref{tab:num_of_leaves}. The number of leaves we add depends on the depth, but we do not add more than 32 leaves in any case, a relatively small number compared to the existing number of leaves. The results are also presented in Table \ref{tab:MNIST}.
We note that growing MNIST trees does not add much value to the full trees, since they are already very highly performant.

We also compare our approach with other tree models, namely Decision Trees, Random Forests, and XGBoost Classifiers. We train one model for each category and select the best hyperparameters using 5-fold cross validation. The results are presented in Table \ref{tab:mnist_other_methods}, where we observe that our approach outperforms the rest of the tree methods. XGBoost Classifier is better than both Random Forest and CART, but although our approach involves a single tree, even shallower Hyperplane GSTs demonstrate a better performance than XGBoost. 

\begin{table}[!ht]\setlength\extrarowheight{1.7pt}
\centering
\begin{tabular}{|c|c|c|c|}
\hline
Model & Best model accuracy (\%) \\
\hline
Hyperplane GST & $98.61$ \\\hline
Convolutional GST & $\mathbf{98.87} $ \\ \hline
CART &$86.00$ \\ \hline
Random Forest  &$94.18$ \\ \hline
XGBoost  & $97.74$   \\ 
\hline
\end{tabular}
\caption{MNIST comparison between different tree methods. The best model is selected based on accuracy on validation set.}\label{tab:mnist_other_methods}%
\end{table}

\subsection{Image Data}

Classification tasks for image data are more challenging compared to tabular data, since images contain more complex information that is harder to extract using simple Hyperplane GSTs. The struggle of Hyperplane GSTs to capture this information lead us to develop Convolutional Trees (Section \ref{subsec:conv trees}) in which the more sophisticated operations at each node increase the modeling power of the tree and present a novel way of computing splits. This modeling power is mostly evident in the case of the CIFAR-10 dataset.  

In this section, we explore image datasets, and we report the performance of both Hyperplane and Convolutional GSTs. The complexity of handling images, primarily because they result in a larger dataset size, and the increased number of parameters of Convolutional Trees, result in longer training times of full trees. That is why in this section we discuss the importance and the effect of growing the trees, as described in Section \ref{deeptree-algo}.


\par\noindent \\
\textbf{Fashion MNIST.} Fashion MNIST is a dataset of clothing images with a training set of 60,000 samples and a test set of 10,000 samples. Each sample is a 28x28 grayscale image that is labeled as one of 10 classes \cite{fashion-mnist}. Intended to serve as a direct drop-in replacement for the original MNIST dataset, it is a more challenging dataset and is also generally well-studied as a benchmark dataset. The state of the art out-of-sample accuracy is 96.91\%, achieved by using a finetuning method called Differential Architecture Search (DARTS) \cite{fashion-mnist-darts}, with other deep neural network models hovering in the 92-94\% accuracy range \cite{fmnist-bench1},\cite{fmnist-bench2}, \cite{fmnist-bench3}, \cite{fmnist-bench4}. 


The method of training follows the same pattern as with MNIST. We finetune hyperparameters by using the same method as with MNIST previously, utilizing trees of depth 4 for this purpose. 

\begin{table}[!ht]\setlength\extrarowheight{1.7pt}
\centering
\begin{tabular}{|c|c|c|c|c|c|}
\hline
Type & Depth & 
\begin{tabular}{@{}c@{}} Average Accuracy \\ (Full) (\%)  \end{tabular} &
\begin{tabular}{@{}c@{}} Average Accuracy \\ (Grown) (\%)  \end{tabular} \\
\hline
Hyperplane & $4$ & $89.51 (\pm 0.25)$  & $89.71 (\pm 0.14)$ \\
& $6$     & $89.78 (\pm 0.25)$ & $90.03 (\pm 0.19)$   \\
& $8$    & $90.04 (\pm 0.12)$ & $90.34 (\pm 0.15)$  \\
& $10$   & $\mathbf{90.11 (\pm 0.25)}$ & $\mathbf{90.34 (\pm 0.12)}$ \\ \hline

Convolutional & $4$ & $90.40 (\pm 0.28)$  & $90.75 (\pm 0.24)$\\
& $6$     & $91.59 (\pm 0.15)$ & $91.74 (\pm 0.20)$ \\
& $8$    & $91.69 (\pm 0.18)$ & $\mathbf{92.06 (\pm 0.22)}$ \\
& $10$   & $\mathbf{91.81 (\pm 0.29)}$ & $91.99 (\pm 0.28)$ \\ \hline

\end{tabular}
\caption{Accuracy of Hyperplane and Convolutional GSTs (Full and Grown) on Fashion MNIST. Average accuracy and standard deviation are calculated across 5 different trees.}\label{tab:fashion-mnist}%
\end{table}

We report the average out-of-sample accuracy for all depths of full and grown trees in Table \ref{tab:fashion-mnist}. Given the fact that our dataset is balanced, accuracy is a valid evaluation metric for our purposes.

\begin{figure}
    \centering 
\begin{subfigure}{0.70\textwidth}
         \includegraphics[width=\textwidth]{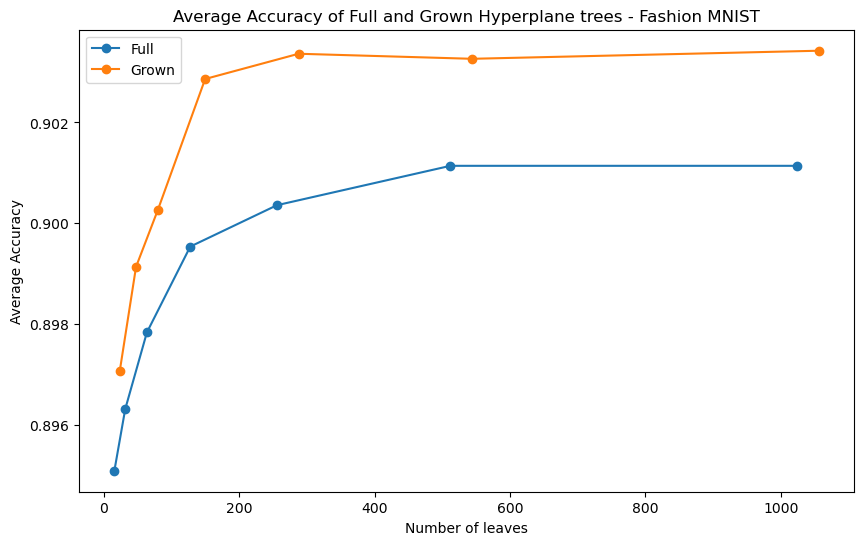}
         \caption{Hyperplane Trees}
         \label{fig:res_fashionMNIST_hyperplane}
\end{subfigure}\hfill
\begin{subfigure}{0.7\textwidth}
         \includegraphics[width=\textwidth]{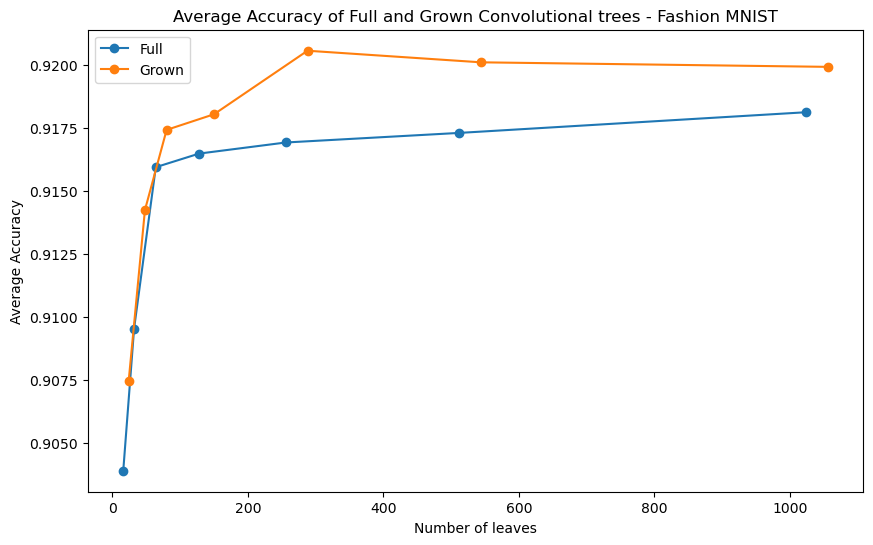}
         \caption{Convolutional Trees}
         \label{fig:res_fashionMNIST_conv}
\end{subfigure}\hfill 
 \caption{Results for Fashion MNIST GSTs. We observe that by adding a small number of leaves to a full a tree can improve the out-of-sample performance. For example, for Hyperplane Trees, a tree with less than 200 leaves performs better than a depth-10 Full Tree, with 1024 leaves.}
 \label{fig:res_fashionMNIST}
\end{figure}

We find that similarly to MNIST, increasing the depth of full trees slightly but consistently improves performance in both the Hyperplane and the Convolutional cases. This indicates that evan shallow trees can capture the images' complexity fairly effectively. Since this dataset is more complicated than MNIST, we observe that Convolutional GSTs outperform Hyperplane GSTs. Growing the trees is also beneficial. We observe that by just adding a small number of leaves, we improve on average over full trees of greater depth, which indicates that we do not necessarily need deep, complete trees to achieve good results, and that extended smaller trees can have an edge. These results are evident in Figure \ref{fig:res_fashionMNIST}.

We also compare the performance of our models to Decision Trees, Random Forests and XGBoost, which we present in Table \ref{tab:fmnist_other_methods}. All of the models are trained on the same training set, similar to the process we follow for MNIST, and we perform hyperparameter tuning using Grid Search. We observe that our Convolutional GSTs outperform the rest of the tree methods, followed by XGBoost, with Hyperplane GSTs achieving performance comparable to XGBoost.

\begin{table}[!ht]\setlength\extrarowheight{1.7pt}
\centering
\begin{tabular}{|c|c|c|c|}
\hline
Model & Best model accuracy (\%) \\
\hline
Hyperplane GST & $90.55$ \\\hline
Convolutional GST & $\mathbf{92.21}$ \\ \hline
Decision Tree &$79.97$ \\ \hline
Random Forest  &$84.42$ \\ \hline
XGBoost  & $90.81$   \\ 
\hline
\end{tabular}
\caption{Fashion MNIST comparison between different tree methods. The best model is selected based on accuracy on validation set.}\label{tab:fmnist_other_methods}%
\end{table}

\par\noindent \\
\textbf{CIFAR-10.} CIFAR-10 is an image dataset, which is widely used in existing literature (\cite{Krizhevsky2009LearningML}). It contains images from 10 different classes, namely airplanes, automobiles, birds, cats, deers, dogs, frogs, horses, ships and trucks. The main difficulty regarding this dataset is the low image quality (32x32x3 pixels), which makes the pattern recognition inside the image hard, in some cases even for the human eye. The full training dataset consists of 50,000 images, with 5,000 images per class, so the dataset is balanced. The test set consists of 10,000 images, and each class is represented by 1,000 images. 

The state of the art out-of-sample accuracy for this dataset is 99.5 \% (\cite{dosovitskiy2021image}), achieved by using a pretrained vision transformer architecture, where the input sequences can be either the raw images, or feature maps extracted from a Convolutional Neural Network. 

Following the same approach as in the experiments previously mentioned, and given the size of the dataset and the images, we first experiment with depth-4 full trees (Hyperplane and Convolutional GSTs). We use the hyperparameter search approach described in Section \ref{subsec:training-methodology} and specify the best hyperparameters, using a 3-fold cross validation approach to calculate the average validation loss and 100 random runs to specify the optimal hyperparameters. We then use these hyperparameters to train 5 trees of depths 4-10, and we present the average accuracy on the test set in Table \ref{tab:CIFAR-10} for selected depths. Since we deal with a multiclass classification problem with a balanced dataset, the accuracy metric score is relevant here.

\begin{table}[!ht]\setlength\extrarowheight{1.7pt}
\centering
\begin{tabular}{|c|c|c|c|c|c|}
\hline
Type & Depth & 
\begin{tabular}{@{}c@{}} Average Accuracy \\ (Full) (\%)  \end{tabular} &
\begin{tabular}{@{}c@{}} Average Accuracy \\ (Grown) (\%)  \end{tabular} \\
\hline
Hyperplane & $4$ & $47.18 (\pm 0.90)$  & $48.92 (\pm 0.32)$  \\
 & $6$     & $51.13 (\pm 0.47)$  & $52.20 (\pm 0.34)$  \\
& $8$    & $52.02 (\pm 0.44)$  & $53.12 (\pm 0.13)$ \\
& $10$   & $\mathbf{52.79 (\pm 0.41)}$   &$\mathbf{53.43 (\pm 0.21)}$ \\ \hline

Convolutional & $4$ & $58.05 (\pm 0.43)$  & $60.82 (\pm 0.51)$  \\
 & $6$    &  $64.56 (\pm 0.60)$ & $65.74 (\pm 0.84)$ \\
 & $8$    &  $66.93 (\pm 0.59)$ & $67.86 (\pm 0.43)$ \\
 & $10$   & $\mathbf{67.74 (\pm 0.58)}$  &$\mathbf{68.3 (\pm 0.24)}$  \\
\hline
\end{tabular}
\caption{CIFAR-10 Full \& Grown GST accuracy (Hyperplane and Convolutional). Average accuracy and standard deviation are calculated across 5 different trees.}\label{tab:CIFAR-10}%
\end{table}

We observe that Hyperplane GSTs are not able to capture the complexity of CIFAR-10 images, and thus only predict correctly around 50\% of the time. Increasing their depth offers a measurable improvement, but overall they are not competitive. Convolutional GSTs improve significantly over Hyperplane GSTs and the difference in performance increases as the depth of the trees increases too. Creating deeper Convolutional Trees is clearly beneficial, especially when the depth grows from 4 to 6. Overall, we note that the improvement in performance is monotonically increasing, namely the deeper the tree the higher the out-of-sample accuracy, with sizeable added value compared to the datasets we have discussed in the previous Sections. 

We also demonstrate the effect of growing the trees using the technique described in Section \ref{deeptree-algo} by training 5 trees for each depth and then growing each by adding a small number of leaves. We present the average results of these experiments in Table \ref{tab:CIFAR-10}, where we clearly see that growing full trees enhances performance, regardless of how shallow the initial tree is. 

\begin{figure}
    \centering 
\begin{subfigure}{0.7\textwidth}
         \includegraphics[width=\textwidth]{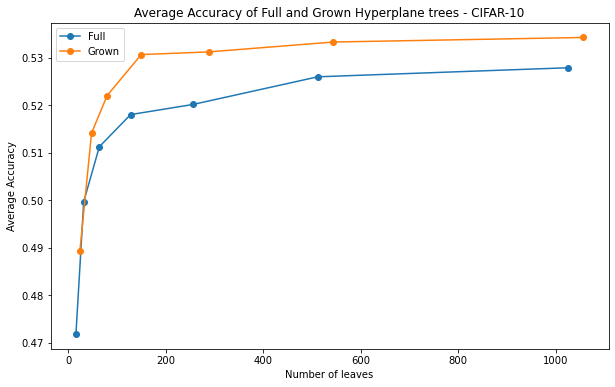}
         \caption{Hyperplane Trees}
         \label{fig:res_cifar_hyperplane}
\end{subfigure}\hfill 
\begin{subfigure}{0.7\textwidth}
         \includegraphics[width=\textwidth]{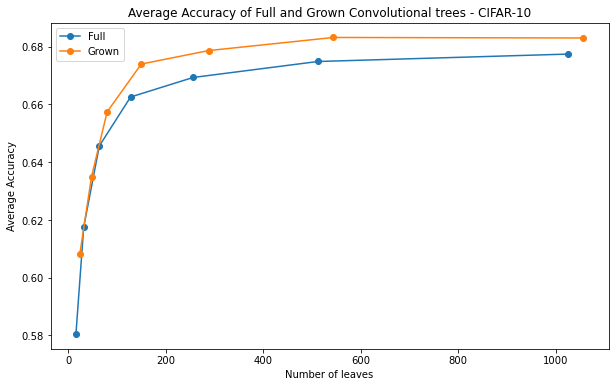}
         \caption{Convolutional Trees}
         \label{fig:res_cifar_conv}
\end{subfigure}\hfill 
 \caption{Results for CIFAR-10 GSTs. We observe that grown trees offer a considerable performance enhancement. For example, in the case of Convolutional Trees, we can grow a depth-7 tree and achieve average accuracy comparable to a full depth-10 tree.}
 \label{fig:res_cifar}
\end{figure}

The true power of our growing technique stems from the creation of trees by adding a small number of leaves, that even outperform full trees of greater depth, as illustrated in Figure \ref{fig:res_cifar}. This means that by starting with a relatively shallow tree and by growing it, we can obtain a more performant tree with less time and memory requirements, compared to a deeper full tree. 

Finally, we compare our best model to other tree methods by training Decision Trees, Random Forests and XGBoost Classifiers and selecting the best based on Grid Search with 5-fold cross validation on the training set. We evaluate their performance on the test set and compare it to our best Hyperplane Tree and Convolutional GST, with results in Table \ref{tab:cifar_other_methods}, where we observe that Convolutional GSTs significantly outperform the rest of the tree methods, many of which are ensembles whereas our models are single trees.


\begin{table}[!ht]\setlength\extrarowheight{1.7pt}
\centering
\begin{tabular}{|c|c|c|c|}
\hline
Model & Best model accuracy (\%) \\
\hline
Hyperplane GST & $53.29$ \\\hline
Convolutional GST & $\mathbf{68.64}$ \\ \hline
CART &$30.46$ \\ \hline
Random Forest  &$43.03$ \\ \hline
XGBoost  & $55.41$   \\ 
\hline
\end{tabular}
\caption{CIFAR-10 comparison between different tree methods. The best model is selected based on accuracy on validation set.}\label{tab:cifar_other_methods}%
\end{table}

Although Convolutional GSTs offer substantial improvement compared to Hyperplane GSTs and the rest of the tree methods, their performance still cannot be compared with deep architectures, like transformers. A factor that increases the problem's complexity is the number of classes. In order to explore how performant Convolutional GSTs are in a binary classification setting, we train one tree per class, so that they learn to discriminate between this specific class and the rest. 

For the binary classification tree that is trained to discriminate between class $n$, $0 \leq n \leq 9$, and the rest of the classes, we use the convention that samples of class $n$ are treated as class 1 samples, and the rest as class 0. 
Since now for each tree we have 5,000 samples in the training set that belong to class 1 and 45,000 that belong to class 0, we randomly select 555 samples from each of the remaining 9 classes, which will be used as the training samples of class 0, so that our model is trained on a balanced dataset. For each class, we obtain 20 models of depth 10, each trained with a random selection of training samples. We present the results of the experiments in Table \ref{tab:bin_cifar} on the test set. Here, the dataset has become imbalanced and we deal with binary classification, so the most appropriate metric to use is AUC. 

\begin{table}[!ht]\setlength\extrarowheight{1.7pt}
\centering
\begin{tabular}{|c|c|c|c|}
\hline
Primary class & Average AUC (\%) \\
\hline
Airplane & $93.45  (\pm 0.17) $ \\ \hline
Automobile & $95.62 (\pm 0.17)$ \\ \hline
Bird & $85.70 (\pm 0.56)$  \\ \hline
Cat &$84.75 (\pm 0.39)$    \\ \hline
Deer&$88.70 (\pm 0.55)$   \\ \hline
Dog & $89.76 (\pm 0.27)$   \\ \hline
Frog & $95.52 (\pm 0.14)$  \\ \hline
Horse & $93.77 (\pm 0.26)$  \\ \hline
Ship & $\mathbf{95.92 (\pm 0.18)}$   \\ \hline
Truck  & $94.00 (\pm 0.33)$    \\
\hline
\end{tabular}
\caption{CIFAR Binary Classification Full Convolutional GSTs. Average AUC is calculated across 20 models.}\label{tab:bin_cifar}%
\end{table}

We observe that the performance of the Convolutional GSTs in the case of the binary classification problem is much better compared to the full 10-class problem and the results are stable, as indicated by the low standard deviation of the AUC. This indicates that a key difficulty of the full problem is its multiclass nature, and that Convolutional GSTs are able to actually discriminate classes in an one-vs-rest set-up. 


\par\noindent \\
\textbf{Celeb-A.} Celeb-A is an image dataset that contains photos of people (mainly their upper bodies and faces) (\cite{liu2015faceattributes}). It contains more than 200,000 images of both males and females, and each image is annotated for multiple features, e.g. male/female, smiling/non-smiling, wearing glasses/non-wearing glasses etc. The task we selected experimenting on is discriminating between males and females.

For the purpose of our experiments, we use 60,000 images, from which 30,000 are pictures of males and 30,000 of females, randomly selected. We then split the dataset into training and test set, which consist of 50,000 and 10,000 images respectively, both balanced in terms of class members. The size of the original images is 218x178 pixels, but for our experiments we downsampled them to 109x89 pixels for scalability reasons and to be able to train deeper trees.

Again, in order to select the best hyperparameters for our models and their training process, we use a 3-fold cross validation approach, we get the average validation loss and we finally utilize it to find the optimal hyperparameters, after 100 random runs.
We then use these hyperparameters to train trees of different depths, in the range of 4-8 and present the results for selected depth values in Table \ref{tab:CELEB-A}. We again calculate the average accuracy of our models, given that our dataset is balanced. We observe that both Hyperplane and Convolutional GSTs are very successful at discriminating between males and females. Convolutional GSTs outperform Hyperplane GSTs, as expected, but their difference in performance is much smaller compared to the case of CIFAR-10.

\begin{table}[!ht]\setlength\extrarowheight{1.7pt}
\centering
\begin{tabular}{|c|c|c|c|c|c|}
\hline
Type & Depth & \begin{tabular}{@{}c@{}} Average Accuracy \\ (Full) (\%)  \end{tabular} &
\begin{tabular}{@{}c@{}} Average Accuracy \\ (Grown) (\%)  \end{tabular} \\
\hline
Hyperplane & $4$ & $94.41 (\pm 0.61)$  & $95.03 (\pm 0.25)$ \\
 & $6$     & $94.62 (\pm 0.27)$  & $\mathbf{95.28 (\pm 0.09)}$  \\
 & $8$   & $\mathbf{94.98 (\pm 0.29)}$ & $95.26(\pm 0.40)$ \\ \hline

Convolutional & $4$ & $96.05 (\pm 0.14)$ & $96.28 (\pm 0.13)$ \\
 & $6$    &  $96.62 (\pm 0.27)$  & $\mathbf{96.69 (\pm 0.15)}$ \\
 & $8$    &  $\mathbf{96.66 (\pm 0.30)}$  &  -\\
\hline
\end{tabular}
\caption{Celeb-A Full \& Grown GSTs accuracy (Hyperplane and Convolutional). Average accuracy and standard deviation are calculated across 5 different trees.}\label{tab:CELEB-A}%
\end{table}

We now demonstrate how growing the trees improves their performance via experiment results of Full and Grown Trees, both Hyperplane and Convolutional, in Table \ref{tab:CELEB-A}. We observe that here, and especially with Hyperplane trees, the grown trees perform comparably or better than full trees of higher depth, indicating that having a much larger number of leaves is not necessary or as beneficial. As with other datasets, the accuracy is monotonically increasing along with the depth of the tree.


Finally, we compare the performance of our models with other tree methods in Table \ref{tab:celeb_other_methods}. For Decision Trees, XGBoost and Random Forests the hyperparameters are selected based on Grid Search, and our best model is considered the one with the highest accuracy on the validation set. We find that all tree methods have a much stronger performance compared to CIFAR-10. Our Convolutional GST results in the best out-of-sample accuracy, followed by XGBoost and then the Hyperplane GST. 

\begin{table}[!ht]\setlength\extrarowheight{1.7pt}
\centering
\begin{tabular}{|c|c|c|c|}
\hline
Model & Best model accuracy (\%) \\
\hline
Hyperplane GST & $95.67$ \\\hline
Convolutional GST & $\mathbf{97.00}$ \\ \hline
CART &$85.18$ \\ \hline
Random Forest  &$92.45$ \\ \hline
XGBoost  & $96.45$   \\ 
\hline
\end{tabular}
\caption{Celeb-A comparison between different tree methods. The best model is selected based on accuracy on validation set.}\label{tab:celeb_other_methods}%
\end{table}

\section{Interpretability of Generalized Soft Trees}\label{sec:interpret}

GSTs provide more insight into the classification process than black box models like Neural Networks. The interpretability advantages of GSTs are three-fold. First, GSTs can be made more interpretable by regularization using a sample penalty, which reduces the number of paths and leaves we need to understand. Second, the input to each node is the original features of the data without nested non-linear transformations. As a result, we can understand how each node partitions the original feature space. For tabular classification tasks, we can directly study the coefficients of the hyperplanes to determine the influence of individual features on the splits. For image classification tasks, since the input to each node is the original image, we can recover meaningful features of the original image used in the splits. Third, GSTs are significantly more shallow in depth than deep neural networks, making it easier to distill and even visualize the full decision-making process of the trees. In this section, we discuss techniques to improve the interpretability of the GSTs, and present approaches to interpret different types of GSTs.

\subsection{Improving Interpretability with Regularization}

\begin{figure}
    \centering 
\begin{subfigure}{0.6\textwidth}
         \includegraphics[width=\textwidth]{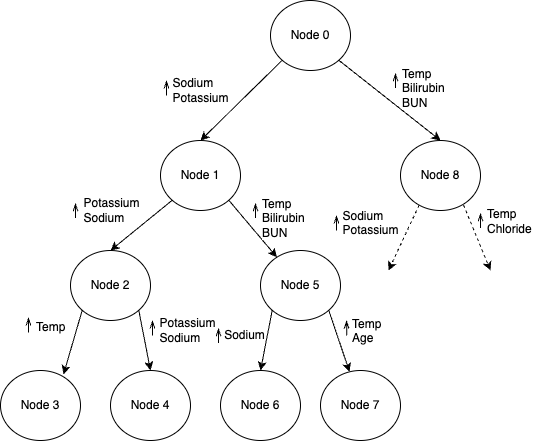}
         \caption{Node splits without sample penalty}
         \label{fig:tree_wo_sp}
\end{subfigure}\hfill 
\begin{subfigure}{0.6\textwidth}
         \includegraphics[width=\textwidth]{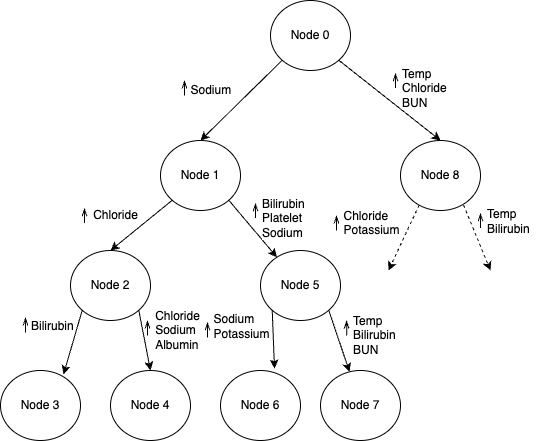}
         \caption{Node splits with sample penalty}
         \label{fig:tree_w_sp}
\end{subfigure}\hfill 
 \caption{Sample node splits for MIMIV-IV depth 4 full tree. We present up to three features with high absolute coefficients for each node. We observe that the tree trained with sample penalty learn different splits at each inner node.}
 \label{fig:tabular_tree_splits}
\end{figure}

In Section \ref{subsec:regularization}, we introduce the notion of a \textbf{sample penalty}, intended to encourage the leaf weights for any given sample to be concentrated across a few leaves, rather than spread across all the leaves. This regularization aids in the interpretability of the final model, as we only need to investigate a smaller number of paths and leaves in the tree to understand the prediction of a given sample, rather than trying to interpret a larger ensemble of leaves, with no distinguishable paths through the tree. As we demonstrate in Section \ref{section:mimic4} and Figure \ref{fig:tabular_tree_splits}, this regularization can lead to more variety in the hyperplane splits, as well as an increase in model performance. 

We define a metric, the $(k, \alpha)$-fraction, which intuitively, measures how concentrated the leaf weights become after training with this regularization. We therefore use the  $(k, \alpha)$-fraction to assess whether the regularization achieves the original goal of concentrating the leaf weights for any given sample, without sacrificing model performance. More rigorously, we define the $(k, \alpha)$-fraction as the percent of validation data such that the top $k$ leaves for that sample, ordered by leaf weight, comprise at least weight $\alpha$, corresponding to $\alpha$-percent of the total weight. This measure is flexible, as the value of $k$ and $\alpha$ may depend on the size of the tree, while also providing a concrete and interpretable measure of how ``spread out'' the weights are for each sample. Using this measure, we evaluate the Convolutional GSTs, trained with and without this sample penalty, at three different sizes: full depth 4, 6, and 8 (corresponding to 16, 64, and 256 leaves, respectively). We report our results for the CIFAR-10 dataset in Table \ref{tab:cifar_sample_penalty}.

\begin{table}[!ht]\setlength\extrarowheight{1.7pt}
\centering
\begin{tabular}{|c|c|c|c|c|}
\hline
Metric/Depth & 4 & 6 & 8\\
\hline
 Best model accuracy (Full) (\%) & $58.62$ & $65.47$ & $67.54$\\
 Best Model accuracy (sample penalty) (\%) & $60.58$ & $65.77$ & $66.14$ \\
 \hline
 (5, 50\%)-recovery (w/o penalty) & 0.1032 & --  & --\\
 (5, 50\%)-recovery (w/ penalty) & 0.8518 & -- & -- \\
 \hline
 (15, 50\%)-recovery(w/o penalty) & -- & 0.0684 & --\\
 (15, 50\%)-recovery(w penalty) & -- & 0.8555 & --\\
 \hline
  (50, 50\%)-recovery(w/o penalty) & -- & -- & 0.4414\\
 (50, 50\%)-recovery(w penalty) & -- & -- & 0.8252\\
 \hline
\end{tabular}
\caption{Comparison of model accuracy and $(k, \alpha)$-fraction of full trees trained with and without the sample-wise penalty on the CIFAR-10 dataset. The $(k, \alpha)$-fraction provides a measure of the concentration of weight among only a few leaves.}\label{tab:cifar_sample_penalty}%
\end{table}
In Table \ref{tab:cifar_sample_penalty}, we select three values for $k$ and $\alpha$ that we found to be appropriate for each model size. The main takeaway from these results is that, especially on smaller trees, we are able to preserve or increase model performance while substantially increasing the concentration of leaf weights. This effect seems to decay as we increase the size of the model, as the depth-8 tree sees both a decrease in accuracy with relatively less improvement in leaf weight concentration. Combining these results with the efficient tree structures that are learned with our DeepTree algorithm, yields a promising direction for developing more interpretable GSTs.

\subsection{Interpreting GSTs via Node Coefficients}\label{section:mimic4}
Hyperplane GSTs can be interpreted by investigating the trained linear coefficients at each of their inner nodes. In this way, we can understand which of the features each node splits on; if the coefficient for a feature is positive, it sends samples with high values for the feature to the right branch of the node and vice versa. When using hyperplane GSTs for image classification tasks, we can visualize filters learned at each node by plotting the coefficients.

\par\noindent \\
\textbf{Hyperplane GSTs for tabular data: MIMIC-IV} \\

Since MIMIC-IV is a purely tabular dataset, we can interpret the GSTs by studying the linear coefficients of the splits. We present sample inner node splits from full depth 4 GST trained with and without sample penalty in Figure \ref{fig:tabular_tree_splits}. For each of the inner nodes presented, we select up to three features that significantly contribute to sending samples to the left/right branch of the node. For example, node 0 in Figure (\ref{fig:tree_wo_sp}) sends samples with high temperature, Bilirubin and BUN values to the right branch and samples with high Sodium and Potassium values to the left branch. In comparing how the inner nodes splits of GSTs grown with (\ref{fig:tree_w_sp}) and without (\ref{fig:tree_wo_sp}) sample penalty, we observe that the inner nodes in the tree trained without sample penalty learn similar splits at each node, whereas the splits in the tree trained with sample penalty are generally different. A potential implication is that a GST trained without sample penalty behaves similarly to an ensemble of generally the same classifier at the leaf nodes. On the other hand, the GST trained with sample penalty can learn different decision rules at each split that can meaningfully split the sample space, which could explain the edge in performance (AUC) compared to trees trained without sample penalty, as observed in Table \ref{tab:mimic_full_tree}. 

\par\noindent \\
\textbf{Hyperplane GSTs for image data: MNIST}\\

\begin{figure}[!ht]
    \centering 
\begin{subfigure}{0.6\textwidth}
         \includegraphics[width=\textwidth]{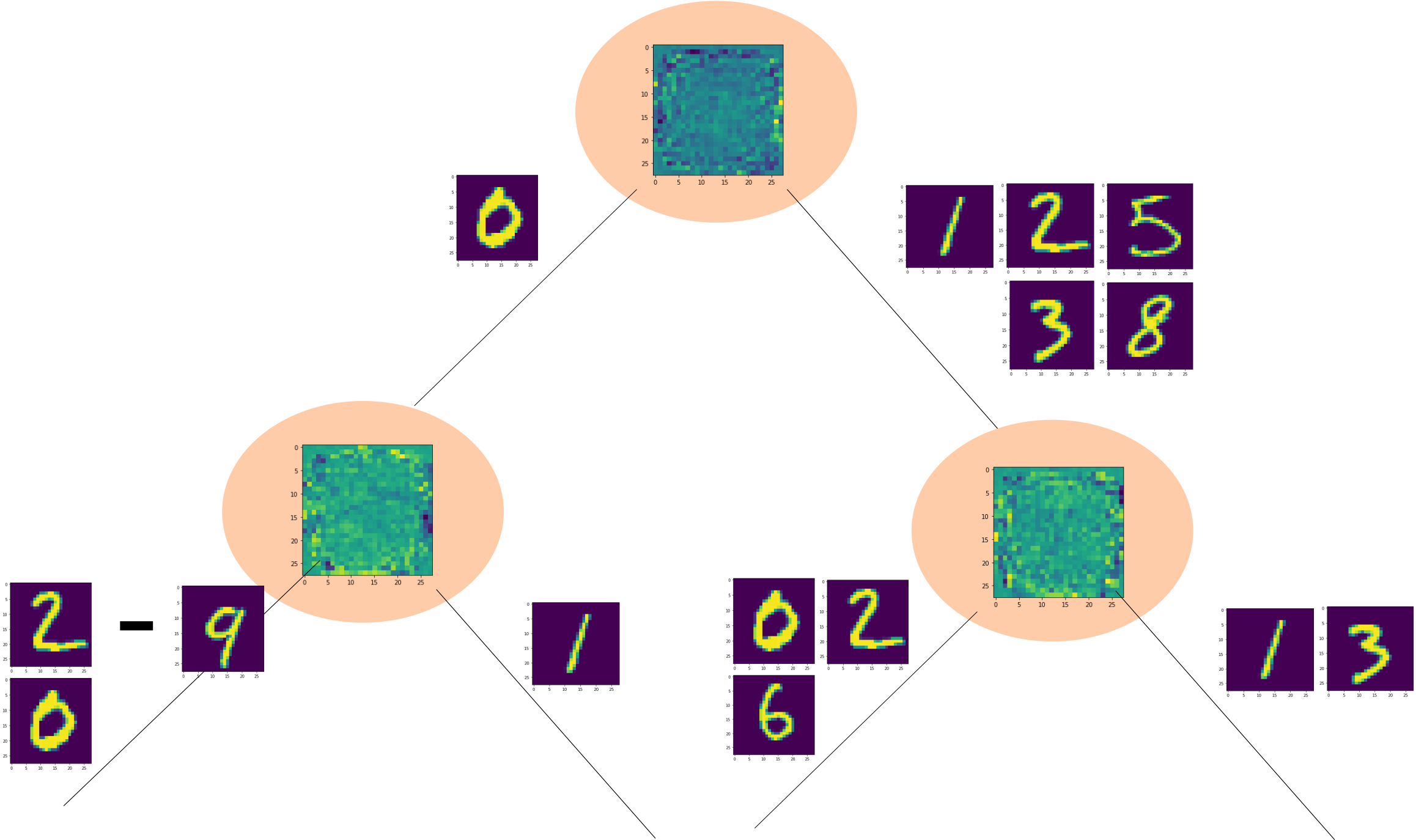}
         \caption{Root node, left and right child.}
         \label{fig:mnist_nodes_0_1_32}
\end{subfigure}\hfill 
\begin{subfigure}{0.6\textwidth}
         \includegraphics[width=\textwidth]{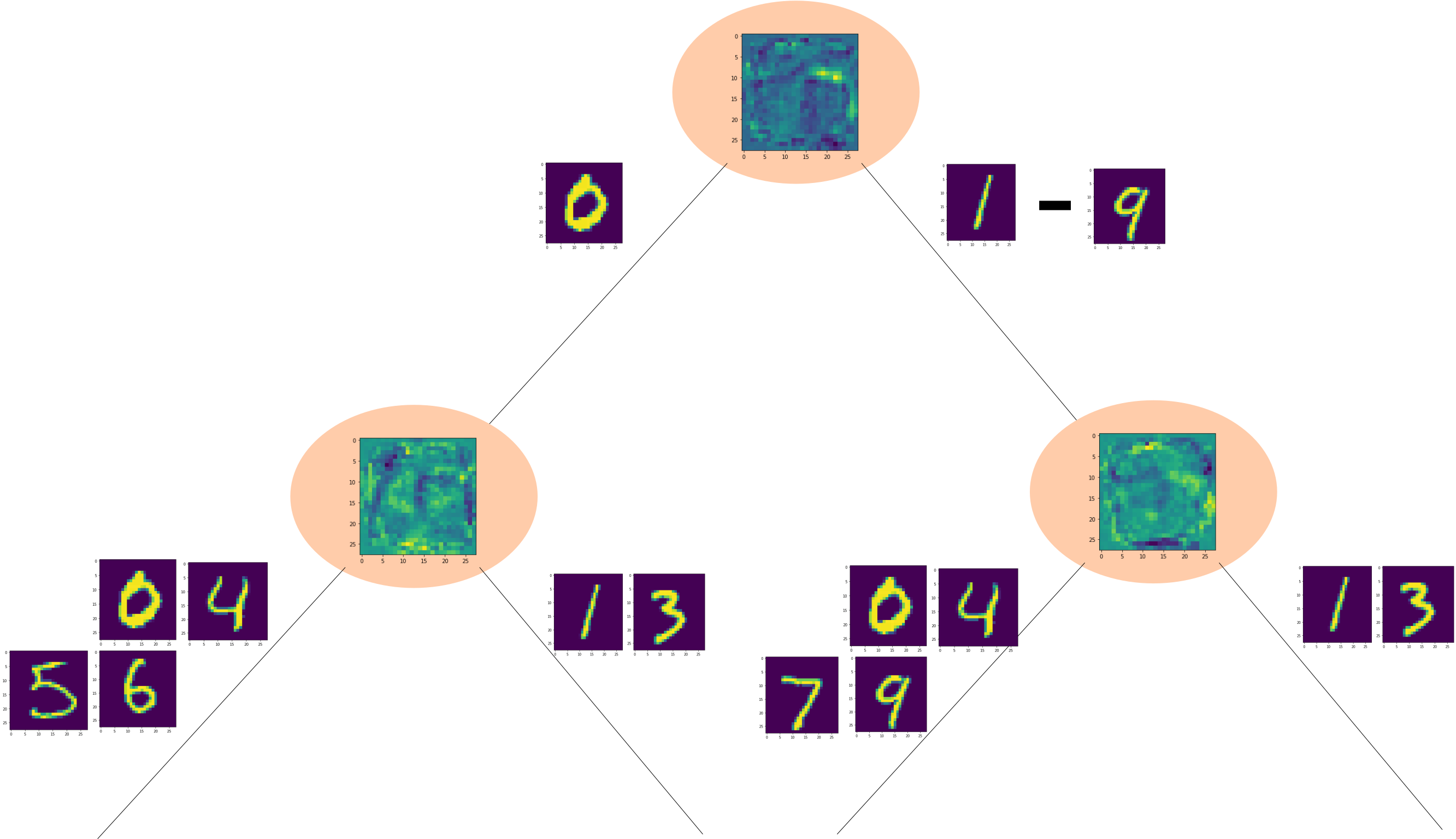}
         \caption{Node of depth 4 and its child nodes.}
         \label{fig:mnist_nodes_26_27_28}
\end{subfigure}\hfill 
\caption{Anatomy of a Hyperplane Tree, trained for the MNIST dataset. The linear coefficients at each split are rearranged and plotted as 28x28 images. We see that 0s and 1s are always assigned to different child nodes, and that larger depth coefficients have more meaningful and complex structure.}
\label{fig:male_female_example_1}
\end{figure}

We can interpret Hyperplane GSTs used for image classification tasks by visualizing the linear coefficients at each split. Here we provide a visualization of these coefficients of our MNIST experiments. We rearrange and plot the coefficients as 28x28 images (the original image shape) and present them for a depth-6 tree in Figure \ref{fig:male_female_example_1}. Figure \ref{fig:mnist_nodes_0_1_32} and \ref{fig:mnist_nodes_26_27_28} visualize the root node and a depth-4 node, respectively, and their respective child nodes.

We observe that the coefficients at small depths do not form specific shapes, but they seem to focus on a ring-like area. In the Figure, we have also included the digits strongly directed to each child (meaning that the vast majority of the samples are assigned to it). We observe that 0s and 1s are always assigned to different child nodes, which is reasonable given their different shape. At a larger depth, the coefficients have a more meaningful and complex structure, giving higher values to specific parts of the input image. We observe that in both child nodes, digits 1 and 3 follow the same path, whereas 0 is again directed to a different child node. These nodes are also the most balanced in terms of separating the different classes.

\par\noindent \\
\textbf{Convolutional GSTs for image data: Celeb-A}\\

Our final method of interpreting Convolutional GSTs is via linear layer coefficients, where we interpret the splits of a Convolutional GST by exploring the coefficients of the linear layer that is applied to the output of each node's convolutional layer. These coefficients multiply each element of the resulting output (feature map) and thus, similar to Hyperplane GSTs, indicate how important each element is in determining the split of each node. 

\begin{figure}
    \centering 
\begin{subfigure}{0.25\textwidth}
  \includegraphics[width=\linewidth]{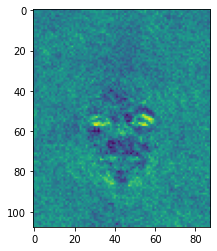}
  \caption{Node 2}
  \label{fig:1}
\end{subfigure}\hfill 
\begin{subfigure}{0.25\textwidth}
  \includegraphics[width=\linewidth]{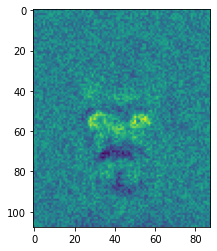}
  \caption{Node 5}
  \label{fig:2}
\end{subfigure}\hfill 
\begin{subfigure}{0.25\textwidth}
  \includegraphics[width=\linewidth]{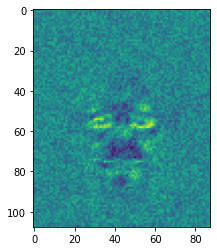}
  \caption{Node 13}
  \label{fig:3}
\end{subfigure}
\medskip
\begin{subfigure}{0.25\textwidth}
  \includegraphics[width=\linewidth]{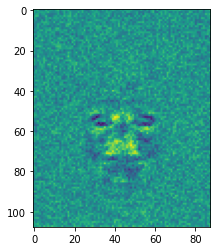}
  \caption{Node 16}
  \label{fig:4}
\end{subfigure}\hfill 
\begin{subfigure}{0.25\textwidth}
  \includegraphics[width=\linewidth]{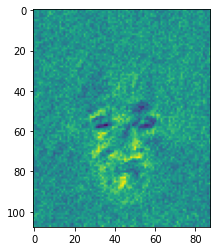}
  \caption{Node 21}
  \label{fig:5}
\end{subfigure}\hfill 
\begin{subfigure}{0.25\textwidth}
  \includegraphics[width=\linewidth]{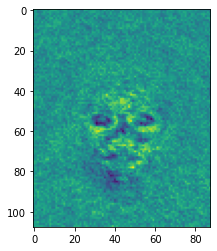}
  \caption{Node 28}
  \label{fig:6}
\end{subfigure}
\caption{Coefficients of the linear layer in different tree nodes. We observe how different nodes highlight different features of the images. These differences cause the top row of nodes to split female and male images to the right and left, respectively, while the bottom row splits female and males to the left and right, respectively.}
\label{fig:images}
\end{figure}

We give an example of this method using the Celeb-A experiment. For the purpose of this section, we train a depth-5 Convolutional GST, using the full size of the dataset images (without downsampling them, to preserve their high resolution). This tree contains $2^5 - 1 = 31$ inner convolutional nodes, and we select, after inspection, 6 of them, whose linear layer coefficients have meaningful shapes that can be seen in Figure \ref{fig:images}. 

Based on the coefficients with the highest values, Figures \ref{fig:1}, \ref{fig:2}, \ref{fig:3} mostly focus on the area of the eyes, as opposed to Figures \ref{fig:4}, \ref{fig:5}, \ref{fig:6}, which focus on the nose and mouth regions. Additionally, based on how many male and female samples are directed to the left and to the right child of each of the nodes, the relative values of the coefficients also inform us how important they are in making classification decisions between males and females in this particular setting. 

In Table \ref{tab:celeb_nodes}, we present the percentage of male and female samples of the test set that are assigned to the right child of each node. We note that due to the softness of the trees, assignment to a node is not binary, so an assignment to the right child means that the weight put on it is greater than the weight on the left child. In nodes 2, 5 and 13, the majority of female samples fall into their right children, whereas male samples mostly fall into the left children. The opposite observation holds for nodes 16, 21 and 28. This suggests that each of these nodes, focusing on a specific subset of the face, groups male and female samples successfully. 

\begin{table}[!ht]\setlength\extrarowheight{1.7pt}
\centering
\begin{tabular}{|c|c|c|c|}
\hline
Node & Male samples (\%) & Female samples (\%) & Face characteristics \\
\hline
2 & $30.98$ & $71.92$ & Eyes \& face shape \\\hline
5 & $29.66$ & $82.86$ & Eyes \& the area between them\\ \hline
13 &$21.86$ & $70.42$ & Eyes \& cheeks \\ \hline
16 &$78.20$ & $23.20$ & Nose\\ \hline
21  & $68.98$ & $30.56$ & Nose \& mouth  \\ \hline
28 & $66.70$ & $25.30$ & Forehead \& area around the eyes\\ 
\hline
\end{tabular}
\caption{Samples of males and females that fall into the right child of particular inner nodes.}\label{tab:celeb_nodes}%
\end{table}

\begin{figure}
    \centering 
\medskip
\begin{subfigure}{0.85\textwidth}
  \includegraphics[width=\linewidth]{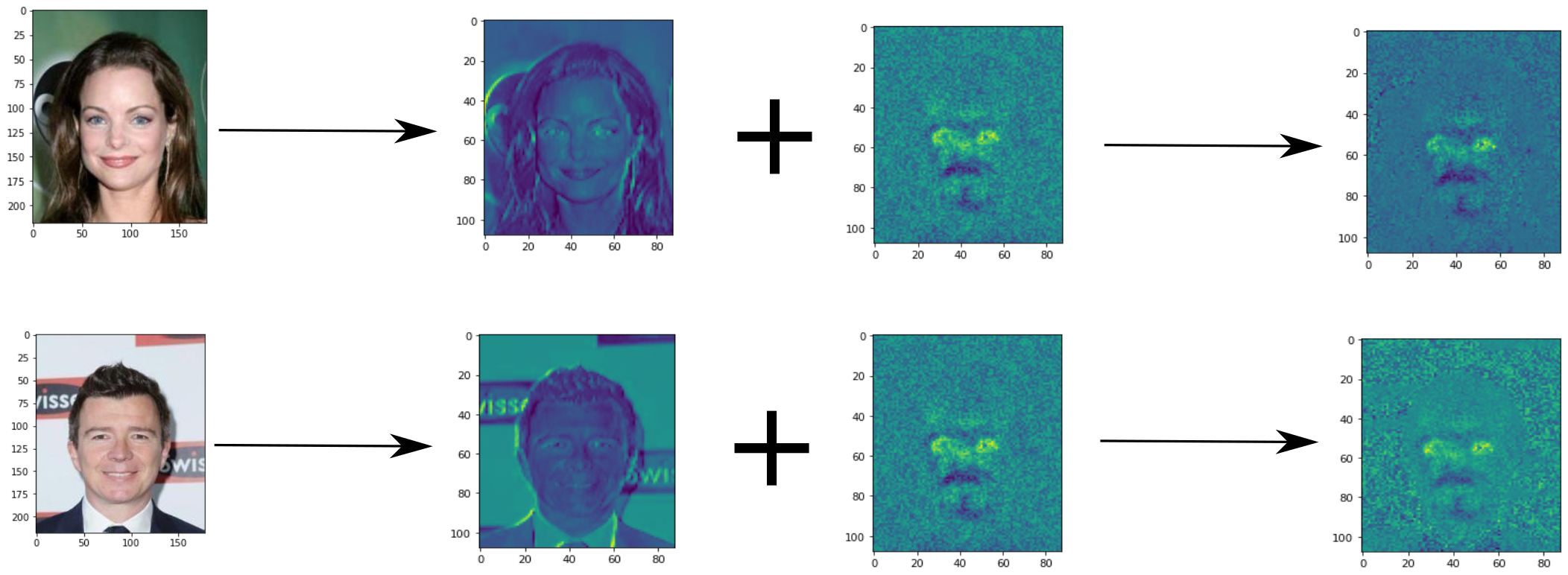}
  \caption{Node 5 - Example.}
  \label{fig:node_5_example}
\end{subfigure}\hfill 
\begin{subfigure}{0.85\textwidth}
  \includegraphics[width=\linewidth]{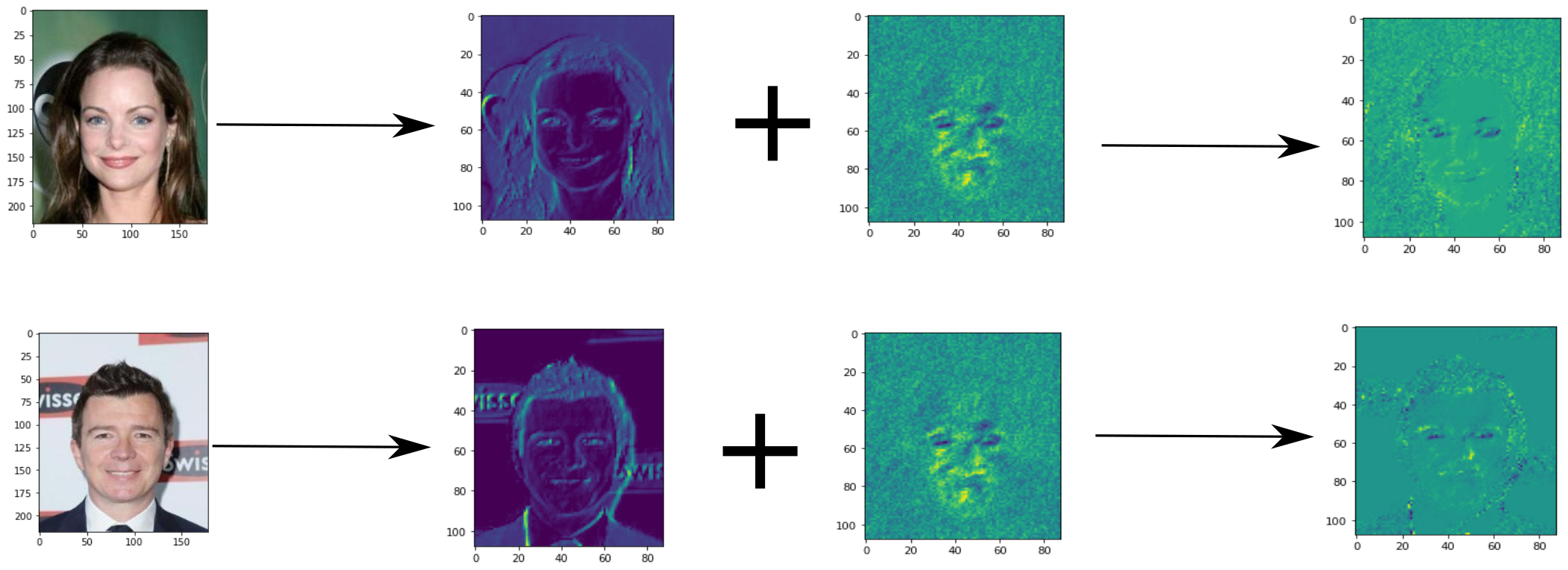}
  \caption{Node 21 - Example.}
  \label{fig:node_21_example}
\end{subfigure}\hfill 
\caption{Examples of convolutional outputs and coefficients of the linear layer for two different nodes. The original image (leftmost) is passed through the convolutional layer (middle images), and then produces the final (rightmost) image after applying the linear layer. The variations between nodes reveal
how they focus on different areas of the image.}
\label{fig:male_female_example_2}
\end{figure}

We conclude this section of interpretability via the linear layer coefficients by presenting how specific input images may be processed inside specific inner nodes and how the coefficients of linear layers are applied in these nodes. We present these examples for inner nodes 5 and 21 in Figures \ref{fig:node_5_example} and \ref{fig:node_21_example}, respectively, and we present the corresponding numeric results in Table \ref{tab:celeb_example_res}, where we confirm our example images are correctly assigned to different child nodes of nodes 5 and 21. 

In Figures \ref{fig:node_5_example} and \ref{fig:node_21_example}, the first (from the left) image is the input image. The second and third images (middle column) result from passing through the convolutional layer of the respective node (convolution operation, ReLU activation function, max pooling). Based on the variation between nodes with regards to their convolutional layer output, we conclude that each node focuses on different areas of the image. The final image (rightmost) is the result of applying the coefficients of the linear layer to the image. The Convolutional GST then compares the sum of the final output components to a threshold to decide the split. 

We observe that the output maps of the two nodes differ significantly; node 5's convolutional layer does not alter the image much, which results in the coefficients prevailing in the final output. On the other hand, node 21's convolutional layer puts more emphasis on specific areas of the image (hair, eyes, facial characteristics lines), so the resulting map after applying the linear coefficients mostly focuses on exactly the intersection of the ``pixels'' the convolutional output and the linear coefficients emphasize.

\begin{table}[!ht]\setlength\extrarowheight{1.7pt}
\centering
\begin{tabular}{|c|c|c|}
\hline
Node & Male example & Female example \\
\hline
5 & $-0.2125$ (Left) & $0.5311$ (Right)\\ \hline
21 &$0.1117$ (Right) & $-0.3869$ (Left)\\ \hline
 
\hline
\end{tabular}
\caption{Resulting output and corresponding split result of the two examples for nodes 5 and 21.}\label{tab:celeb_example_res}%
\end{table}

\subsection{Interpreting GSTs via Feature Visualization}


Feature visualization \cite{feature_vis}  is an established method for interpreting individual nodes and layers within deep neural networks that is applicable to Convolutional GSTs. For each inner node in the tree, we train an image input to both maximize and minimize the activation of that node. Formally, for any inner node $n$, we are solving the optimization problem $\max_{x \in X} \sigma(f_n(x))$, where the input image $x$ is trained using gradient descent. Accordingly, we find the optimal image for minimizing the node activation the same way, by minimizing $\sigma(f_n(x))$. There are many forms of regularization for this feature visualization method, but we find that applying an $L_1$ and $L_2$ penalty on the input image is sufficient to remove most of the noise in the optimized image. 

Unlike deep neural networks, each inner node in our GSTs is a separate function of the input, rather than being nested. These inner nodes contain only one convolutional and linear layer, so this method of feature visualization provides a much more direct interpretation of what has been learned for each split in the tree than this method provides for interpreting what has been learned for a deep neural network. 

\begin{figure}
     \centering
     \includegraphics[width=\textwidth]{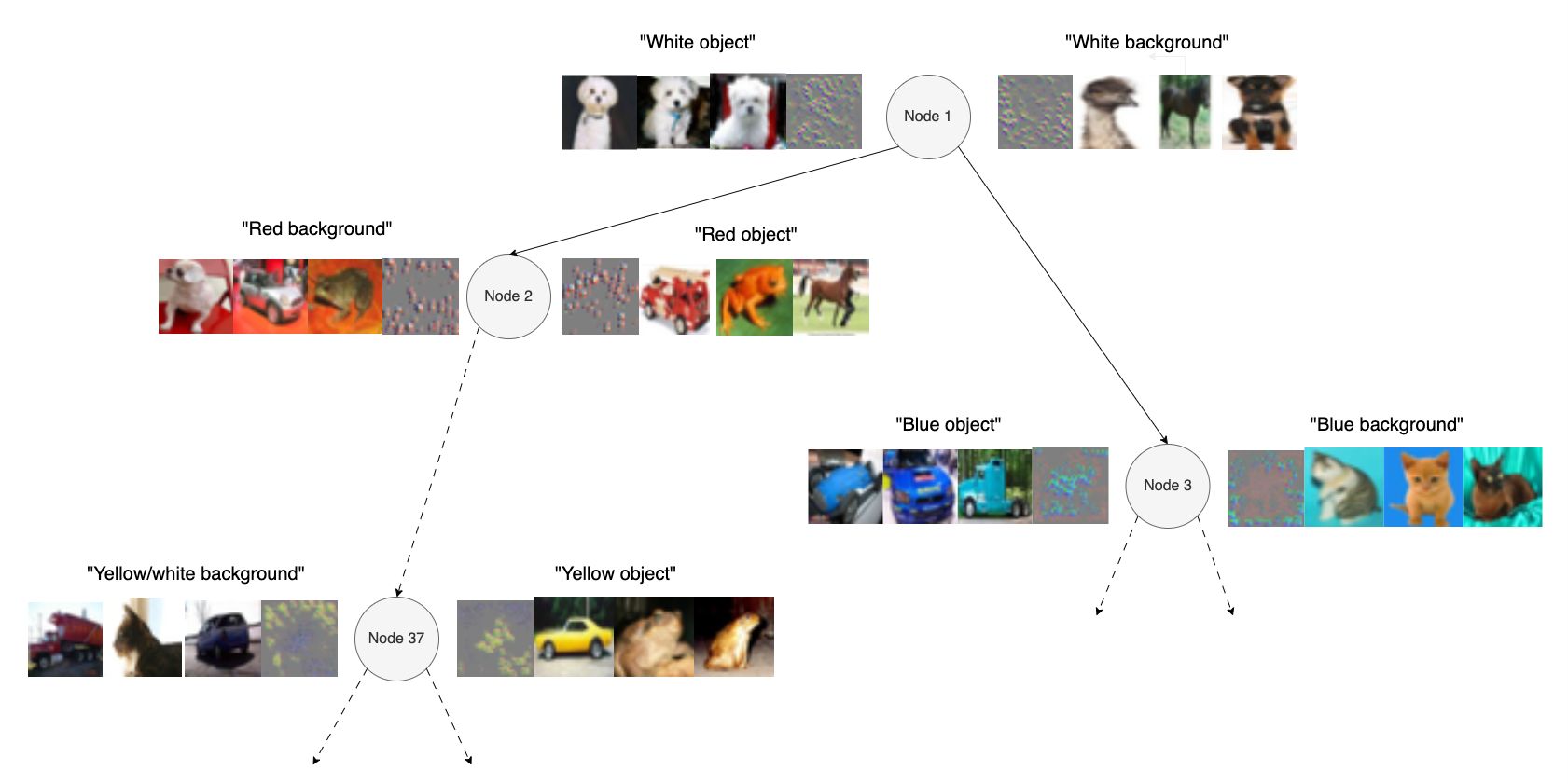}
     \caption{Feature visualizations for a depth-6 tree trained on CIFAR-10. To the right of each node is an image generated to maximize the activation of that node, along with the top-3 real images that maximized the activation. Correspondingly, we present the images that minimize the activation on the left of each node.}
     \label{fig:cifar-feature-vis}
\end{figure}

We present an example of these visualizations in Figure \ref{fig:cifar-feature-vis}, with a depth-6 convolutional tree trained on CIFAR-10 data. For some of the nodes, clear patterns emerge. For example, in Figure \ref{fig:cifar-feature-vis}, we see that Node 3 has clearly learned to split images based on whether they have a blue background, or a blue object. This split is likely meaningful because there are many images of cats and dogs in the CIFAR-10 dataset that are set on blue backgrounds, and blue objects tend to be either cars or trucks. Therefore, splitting on this property is likely an efficient way to distinguish between animals and vehicles. We see similar splits emerge in many nodes in this tree, distinguishing between a certain color as the background or as an object. 

These visualizations demonstrate both the improved interpretability of these trees over neural networks, while also highlighting how they struggle to learn more abstract concepts, such as shapes or patterns, which likely explains the gap in performance between these trees and deep neural networks on challenging image datasets like CIFAR-10.

\section{Conclusions}\label{sec:conclusions}

GSTs are widely applicable and provide a new perspective on existing tree methods. Both Hyperplane Trees and Convolutional Trees demonstrate tractability, particularly when grown via the DeepTree algorithm. Growing the trees presents a practical and efficient way to create meaningful unbalanced trees, and we demonstrate in specific datasets that the improvement achieved by growing the trees is greater than training deeper full trees, suggesting that the addition of a few nodes is more beneficial than growing full trees in a brute-force manner.

Furthermore, our GSTs showcase promising performance in a variety of data modalities, including both structured (tabular) and unstructured (image) datasets; in experiments on benchmark datasets, GSTs outperform other available tree methods and in some cases have comparable performance to state-of-the art models, which is largely due to the fact that they contain more complex functions at their split nodes. In particular, both Hyperplane and Convolutional GSTs are very successful in experiments with MNIST, Fashion MNIST and Celeb-A, whereas Convolutional GSTs capture the more complicated nature of CIFAR-10 better than Hyperplane GSTs, although they are still not comparable to deep architectures. We also demonstrate how Convolutional GSTs have a significantly improved performance when the CIFAR-10 classification problem is converted into a binary one.

Finally, we manage to preserve some interpretability with our GSTs. Though they are not as straightforwardly interpretable as traditional Decision Trees, our GSTs provide more meaning than non-interpretable traditional Deep Learning models. In our exploration on interpretability in various experimental datasets, we have demonstrated interesting findings and insights on how GSTs operate.

Overall, GSTs introduce a new way of viewing trees; they demonstrate tractability as a result of borrowing methods from the Deep Learning field (convolutional layers, backpropagation) and growing via the DeepTree algorithm, they produce strong results on benchmark datasets with a variety of data modalities, and they successfully preserve some of the interpretability characteristics of trees. 

\section*{Funding}
This work was conducted by the authors while they were affiliated with the Massachusetts Institute of Technology.

\section*{Ethics declarations}
\subsection*{Conflicts of interest/Competing interests}
Not applicable.
\subsection*{Ethics approval}
Not applicable.
\subsection*{Consent to participate}
Not applicable.
\subsection*{Consent for publication}
Not applicable.
\subsection*{Availability of data and material}
All datasets used are publicly available.

\subsection*{Code availability}
The code implementing the experiments is available upon request. 

\section*{Authors' contributions}
Dimitris Bertsimas conceived, supported, and supervised the presented work and edited the paper. Matthew Peroni developed the DeepTree Algorithm. Matthew Peroni and Vasiliki Stoumpou implemented the pipeline and the code. Jiayi Gu, Matthew Peroni and Vasiliki Stoumpou ran the experiments, conducted the corresponding results' analysis and worked on the interpretability insights. Lisa Everest, Jiayi Gu, Matthew Peroni and Vasiliki Stoumpou co-wrote the paper.

\bibliographystyle{alpha}
\bibliography{sn-bibliography}

\end{document}